\pdfoutput=1

\documentclass[11pt]{article}
\usepackage{tikz}

\newcommand{\cmark}{\ding{51}}  
\newcommand{\xmark}{\ding{55}}  
\renewcommand{\cmark}{\textcolor{green}{\ding{51}}}
\renewcommand{\xmark}{\textcolor{red}{\ding{55}}}
\definecolor{codeblue}{RGB}{0, 0, 255}
\definecolor{lightgray}{RGB}{245, 245, 245}  

\newcommand{\halfcheck}{%
  \begin{tikzpicture}[baseline=-0.5ex]
    \node[text=orange, scale=1] at (0,0) {\ding{51}};
    \draw[line width=1.5pt, orange] (-0.25em,0.4em) -- (0.25em,-0.4em);
  \end{tikzpicture}%
}
\usepackage[final]{EMNLP2025}

\usepackage{xcolor}
\usepackage{makecell}
\usepackage{times}
\usepackage{latexsym}

\usepackage[T1]{fontenc}

\usepackage[utf8]{inputenc}

\usepackage{microtype}

\usepackage{inconsolata}

\usepackage{graphicx}
\usepackage{threeparttable}
\usepackage{amssymb}  
\usepackage{pifont}   
\usepackage{multirow}
\usepackage{booktabs}  
\usepackage{colortbl}
\usepackage[most]{tcolorbox}  
\usepackage{enumitem}
\usepackage{longtable}
\usepackage{cuted}
\usepackage{multicol}
\usepackage{mdframed}
\usepackage{parallel}
\usepackage{supertabular}
\usepackage{placeins}
\usepackage{CJK}
\usepackage{float} 
\usepackage{subcaption}

\title{\textsc{RETAIL}: Towards Real-world Travel Planning for\\ Large Language Models}

\author{Bin Deng$^{1\dagger}$ \thanks{$^\dagger$ Equal Contributions. Work done during the first author's internship at Beihang University.}, Yizhe Feng$^{1\dagger}$ , Zeming Liu$^{1\ddagger}$  \thanks{$^\ddagger$ Corresponding Author}, Qing Wei$^{2}$,\\ \bf   Xiangrong Zhu$^{2}$, Shuai Chen$^{2}$, Yuanfang Guo$^{1}$, Yunhong Wang$^{1}$\\
  $^{1}$School of Computer Science and Engineering, Beihang University, Beijing, China, \\
  $^{2}$Meituan Inc., China, \\
  \texttt{db1124@buaa.edu.cn, sy2306328@buaa.edu.cn,  zmliu@buaa.edu.cn} }\date{}
\makeatletter
\def\thanks#1{\protected@xdef\@thanks{\@thanks
        \protect\footnotetext{#1}}}
\makeatother

\begin{document}
\maketitle
\begin{abstract}
Although large language models have enhanced automated travel planning abilities, current systems remain misaligned with real-world scenarios. First, they assume users provide explicit queries, while in reality requirements are often implicit. Second, existing solutions ignore diverse environmental factors and user preferences, limiting the feasibility of plans. Third, systems can only generate plans with basic POI arrangements, failing to provide all-in-one plans with rich details. To mitigate these challenges, we construct a novel dataset \textbf{RETAIL}, which supports decision-making for implicit queries while covering explicit queries, both with and without revision needs. It also enables environmental awareness to ensure plan feasibility under real-world scenarios, while incorporating detailed POI information for all-in-one travel plans. Furthermore, we propose a topic-guided multi-agent framework, termed TGMA. Our experiments reveal that even the strongest existing model achieves merely a 1.0\% pass rate, indicating real-world travel planning remains extremely challenging. In contrast, TGMA demonstrates substantially improved performance 2.72\%, offering promising directions for real-world travel planning.
\footnote{Code and data will be made publicly available.}
\end{abstract}

\section{Introduction}

\begin{figure*}[t]
\centering
\includegraphics[width=\textwidth]{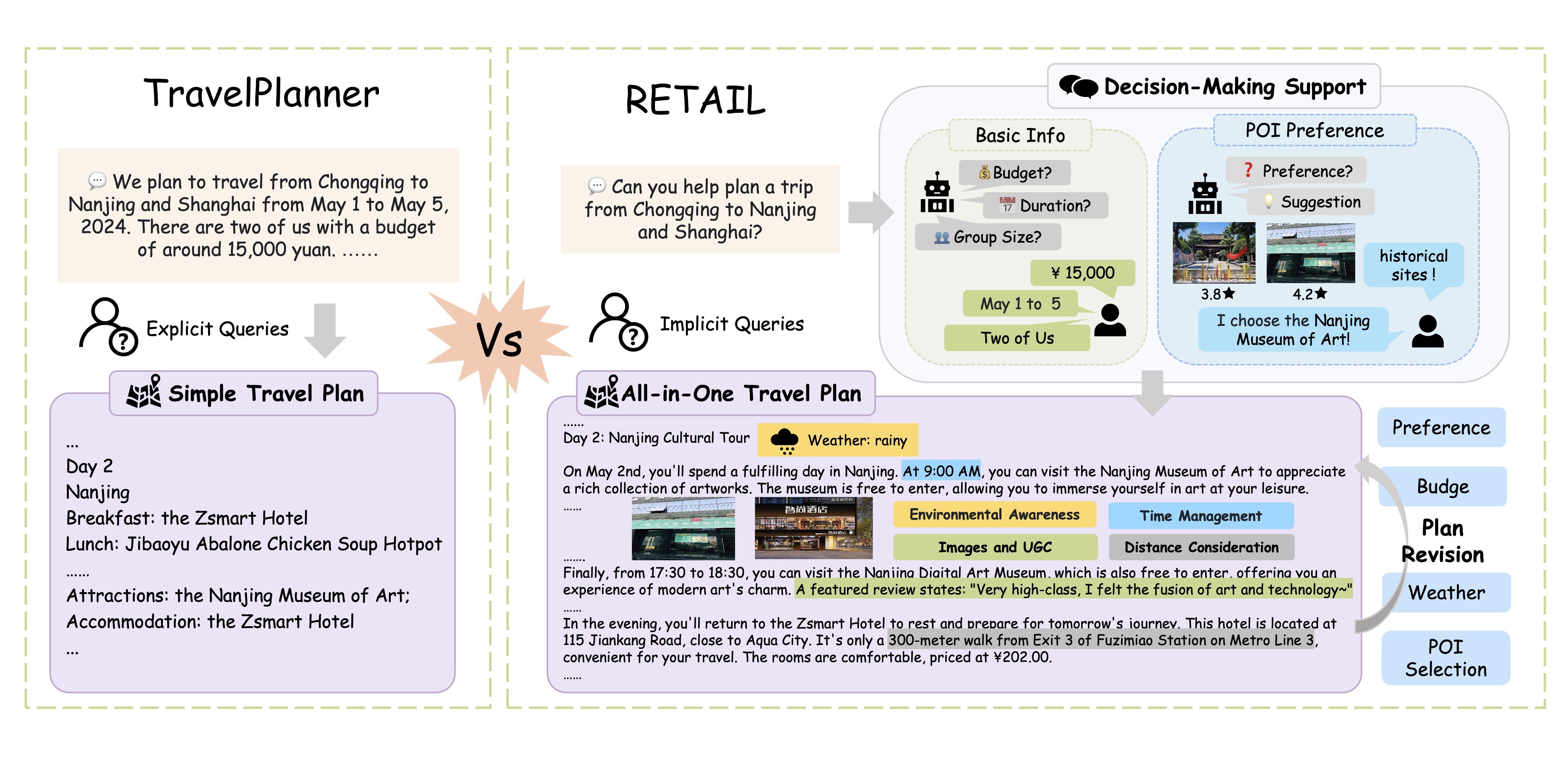}
\caption{Comparison between TravelPlanner~\cite{xie2024travelplanner} and RETAIL, highlighting our UGC-enriched decision-making support for implicit queries, environmental awareness, and all-in-one travel plans with rich details.}
\label{running_example}
\end{figure*}

Early tourism planning systems focused on automating travel plan generation based on user queries~\cite{berka2004designing,lucas2013hybrid}, primarily simplifying information search and basic planning processes. Recent advances in large language models~\cite{touvron2023llama, jiang2023mistral} have enabled more sophisticated approaches that consider user preferences and commonsense knowledge~\cite{xie2024travelplanner,chen2024travelagent,tang2024itinera}

However, existing travel planning research remains misaligned with real-world complexities. First, current approaches~\cite{xie2024travelplanner,singh2024personal} predominantly rely on the assumption that users have explicit intentions when formulating plans, which limits their practical utility. Second, these methods~\cite{chen2024travelagent,zhang2024ask} ignore the variability of user needs and environmental conditions, thus lacking the ability to dynamically adjust plans. Third, while existing work focuses primarily on basic information\cite{xie2024travelplanner, singh2024personal}, it overlooks critical details such as timing, images, ticket pricing, UGC-enriched room info that are essential for real-world practical plans.
As shown on the left side of Figure \ref{running_example}, they typically require users to predefine precise details such as duration, budget, and POI preference. However, this does not align with reality due to the casual and imprecise nature of tourists' speech. Moreover, they also fail to consider dynamic user needs and complex environmental conditions(e.g., rainy days). Additionally, their plans are oversimplified, only including restaurants, attractions, and accommodations, which ultimately leads to a lack of practical feasibility.

To mitigate these challenges, we constructed a \textbf{RE}al-world \textbf{T}ourism \textbf{A}ll-in-one \textbf{I}nteractive p\textbf{L}anning dataset, termed \textbf{RETAIL}. Specifically, it incorporates two complementary mechanisms to support travel planning. The decision-making support process guides users from basic travel requirements to specific preferences through clarification, acknowledging that initial user queries are often implicit and mitigating the first challenge. The plan revision and environmental awareness process can further adjust plans based on changes in user requirements or perceived environmental conditions, which mitigates the second challenge of lacking environmental awareness and plan revision capabilities. It also has all-in-one travel plans that integrate time arrangements, tickets, rooms, cost and etc, ensuring practical feasibility and mitigating the third challenge of lacking comprehensive travel planning. As shown on the right side of Figure \ref{running_example}, RETAIL starts from a implicit query, obtains users' true intentions through a decision-making support process, and then generates all-in-one travel plans, while allowing plan revision in response to environmental changes. 

To improve the ability of real-world travel plan generation, we propose TGMA, a topic-guided multi-agent framework. TGMA features a topic-guided approach and multi-agent collaboration. First, we design a Topic-Guided Interaction Logic that dynamically selects appropriate topics based on dialogue context, enabling natural and expectation-aligned interactions. Second, we develop a multi-agent architecture with three specialized agents: the intent detection agent for extracting accurate user requirements, the overall plan agent for generating structured preliminary plans and the detailed plan agent for producing comprehensive final plans.

This work makes the following contributions:
\begin{itemize}
\item \textbf{We identify challenges in real-world travel planning:} Existing works assume users provide explicit queries rather than implicit requirements. Travel plans need to adapt to dynamic environmental factors and user needs. All-in-one travel plans require integration of various practical details.

\item \textbf{To mitigate these challenges, we present a novel dataset RETAIL,} covering planning from explicit queries and planning after clarifying implicit queries through decision-making support, both with and without revision needs, totaling 10,182 real-world cases. All plans are all-in-one travel itineraries with rich POI information, demonstrating high practical value.

\item \textbf{We design TGMA, a topic-guided multi-agent framework for travel planning.} Experimental results demonstrate TGMA's effectiveness in providing coherent decision support and generating all-in-one travel plans that integrate various practical aspects.
\end{itemize}

\section{Related Work}

\begin{table*}[t]
\small
\setlength{\tabcolsep}{5pt}
\renewcommand{\arraystretch}{1}
\centering
\begin{tabular}{lcccccccccccc}
\toprule
 \multirow{2}{*}{Dataset} & \multicolumn{5}{c}{Real-world Factors} & \multicolumn{4}{c}{Tourism KB} & \multirow{2}{*}{Query} & \multirow{2}{*}{Avg. Turns} & \multirow{2}{*}{Total} \\
\cmidrule(lr){2-6}
\cmidrule(lr){7-10}
& DMS. & EA. & AOP. & PR. & DPI & Att. & Res. & Hot. & Wea. & & &\\
\midrule
ITINERA & \xmark & \xmark & \xmark & \xmark & \xmark & 7,578 & 0 & 0 & 0 & Ex.\&Im. & 1 & N/A \\
TravelPlanner & \xmark & \xmark & \xmark & \xmark & \xmark & 5,302 & 9,551 & 5,047 & 0 & Ex. & 1 & 1,225 \\
Ask-before-Plan & \halfcheck & \xmark & \xmark & \xmark & \xmark & 5,302 & 9,551 & 5,047 & 0 & Ex.\&Im. & 2.8 & 2,000 \\
TravelAgent & \xmark & \xmark & \xmark & \xmark & \halfcheck & N/A & N/A & N/A & N/A & Ex.\&Im. & 1 & 20 \\
ChinaTravel  & \xmark & \xmark & \cmark & \xmark & \halfcheck & 3,413 & 4,655 & 4,124 & 0 & Ex. & 1 & 154 \\
\midrule
\textbf{RETAIL(ours)} & \textbf{\cmark} & \textbf{\cmark} & \textbf{\cmark} & \textbf{\cmark} & \textbf{\cmark} & \textbf{8,229} & \textbf{29,586} & \textbf{22,464} & \textbf{8,784} & Ex.\&Im. & \textbf{4.65} & \textbf{10,182}\\
\bottomrule
\end{tabular}

\caption{Comparison between RETAIL to other datasets, including ITINERA\cite{tang2024itinera},TravelPlanner\cite{xie2024travelplanner},Ask-before-Plan\cite{zhang2024ask}, TravelAgent\cite{chen2024travelagent} and ChinaTravel\cite{shao2024chinatravel}.\textbf{DMS.} for \textbf{UGC-enriched Decision-Making Support}, \textbf{EA.} for \textbf{Environmental Awareness}, \textbf{PR.} for \textbf{Plan Revision}, \textbf{DPI} stands for \textbf{Detailed POI Information, including Images and User-Generated Content like reviews, ratings, and must-visit rankings}, \textbf{AOP} for \textbf{All-in-One Plan}.\textbf{Att.}, \textbf{Res.}, \textbf{Hot.}, \textbf{Wea.} stands for \textbf{Attraction}, \textbf{Restaurant}, \textbf{Hotel}, \textbf{Weather}. \textbf{Ex.}, \textbf{Im.} stands for \textbf{Explicit}, \textbf{Implicit}.
\textbf{N/A} denotes \textbf{Not Applicable}, which indicates that no data is provided. "\xmark" indicates the absence of a dimension, while "\halfcheck" indicates that only part of it is included;for example, \textbf{Ask-before-Plan} only includes \textbf{basic inquiries} without incorporating UGC-enriched recommendations. \textbf{Travel Agent} and \textbf{ChinaTravel} only incorporate \textbf{images} without leveraging UGC content.}

\label{tab:dataset_comparsion}
\end{table*}
\subsection{Traditional Travel Planning}

Traditional travel planning systems have evolved significantly over the past decades, from rule-based approaches to sophisticated recommendation engines~\cite{berka2004designing}. Early systems rely on predefined rules and constraints for basic itinerary generation~\cite{sebastia2009tourism}, later incorporating personalized recommendations~\cite{lucas2013hybrid} and context-aware services~\cite{rodriguez2013gat}. Various optimization techniques have been developed, from multi-day planning~\cite{chen2013automatic} to approaches using clustering and deep learning~\cite{huang2020multi}, as well as weighted mining and ant colony optimization~\cite{yu2022analysis,liang2021improved}. While mobile and context-aware systems emerge as solutions towards more dynamic and personalized travel experiences, the inherent complexity of tourism info and the NP-hard nature of trip design~\cite{gavalas2014survey} remain significant challenges.

These systems primarily formulate travel planning as mathematical optimization problems, excelling at computational scheduling. However, they lack decision-making support to help users clarify implicit requirements.

\subsection{LLM-based Travel Planning}
The emergence of LLMs has revolutionized travel planning through their enhanced capabilities in reasoning~\cite{kojima2022large,wei2022chain}, tool use~\cite{schick2023toolformer, DBLP:conf/iclr/QinLYZYLLCTQZHT24}, and agent frameworks~\cite{shinn2024reflexion,yao2022react}.

Recent research has explored LLM-based travel planning from multiple perspectives. TravelPlanner~\cite{xie2024travelplanner} establishes foundational planning capabilities, further enhanced by TravelPlanner+~\cite{singh2024personal} and ChinaTravel~\cite{shao2024chinatravel} for personalization and evaluation. Advanced planning approaches emerge through ItiNera~\cite{tang2024itinera} and TravelAgent~\cite{chen2024travelagent} for spatial optimization, while TRIP-PAL~\cite{de2024trip} and TTG~\cite{ju2024globe} focus on plan feasibility. For interactive planning, APEC-Travel~\cite{jiang2024towards} and CEP~\cite{zhang2024ask} explore preference extraction through dialogues, while EVOAGENT~\cite{yuan2024evoagent} employs multi-agent collaboration and human-like reasoning~\cite{xie2024human} introduces structured planning strategies. Beyond itinerary generation, LLMs have also demonstrated potential in mobility analysis in transportation~\cite{transportation} and climate analysis~\cite{time}, providing foundational support for environmental awareness in travel planning.

Existing LLM-based approaches focus on simple trip generation and basic requirement clarification. However, they do not achieve UGC-enriched decision-making support, systematic environmental awareness, or all-in-one travel planning driven by rich POI information that our work enables.

\section{RETAIL}
\label{sec:dataset_construct}
To mitigate the limitations in handling implicit requirements, dynamic environmental factors, and all-in-one travel planning, we present RETAIL dataset for real-world travel planning. It supports both direct planning from explicit requirements and clarification through decision-making support, with dynamic plan revision capabilities. A detailed comparison with existing datasets is presented in Table~\ref{tab:dataset_comparsion}.

RETAIL is constructed through three phases: (1) Tourism Knowledge Base Collection; (2) Dataset Construction; and (3) Dataset Annotation.

\subsection{Tourism Knowledge Base Collection}
To support travel plans with rich POI information, we establish a comprehensive tourism database spanning 24 major Chinese cities, integrating 60,279 POIs with essential real-world planning factors such as detailed POI information, transportation data, and weather data. Detailed examples of these information can be found in Appendix~\ref{appendix B}.

\textbf{City Selection.} To reflect real-world tourism patterns and visitor flows across China, we select 24 major Chinese cities that represent diverse travel preferences. These destinations include both metropolitan centers (e.g., Beijing, Shanghai, Guangzhou) and cultural hubs (e.g., Chengdu, Xi'an, Xiamen).

\textbf{POIs Information Collection.} To support all-in-one travel plans, we gather nearby POI information around attractions to ensure distance constraints. Each attraction is linked to 3-5 nearby restaurants and hotels, incorporating essential planning factors: attraction information (categories, tickets, reviews), dining options (costs, ratings), and accommodation choices (locations, room types, guest feedback). POIs data were obtained through web crawling from \textbf{Meituan}\footnote{https://www.meituan.com/}, China's leading O2O service platform, ensuring comprehensive and up-to-date local business information.

\textbf{Weather and Transportation Collection.} To enable environmental awareness, we incorporate weather forecasts and transportation data for travel planning under diverse conditions. The weather data was collected from the \textbf{2345 Weather Service Platform} \footnote{https://tianqi.2345.com/} covering full-year 2024 forecasts, while transportation data were obtained through web crawling from \textbf{Meituan}. These dynamic factors directly impact travel decisions and reflect practical considerations in travel planning.

\subsection{Dataset Construction}
The Dataset construction consists of three processes: (1) Query Construction, (2) Decision-Making Support, and (3) Plan Generation and Adjustments.

\subsubsection{Query Construction}
We categorize queries into explicit and implicit types to simulate real-world travel planning scenarios. \textbf{For explicit queries}, we select one departure city and 2-4 destination cities, identify 8-10 key attractions with dining and accommodation options, and incorporate weather and transportation information through LLM to construct comprehensive, information-rich queries. \textbf{For implicit queries}, based on explicit queries, we develop a 12-field intention slot and strategically remove selected fields to mirror real-world tourist inquiries where preferences are often partially expressed.
    
\subsubsection{Decision-Making Support}

\begin{table*}[!htbp]
\centering
\setlength{\tabcolsep}{10pt}
\renewcommand{\arraystretch}{0.9}
\begin{tabular}{l r l r l r}
\toprule[1.1pt]
\multicolumn{6}{c}{\textbf{Dataset Statistics}} \\
\midrule[0.8pt]
\multicolumn{2}{c}{\textit{Dialogue Level}} & 
\multicolumn{2}{c}{\textit{Utterance Level}} & 
\multicolumn{2}{c}{\textit{Token Level}} \\
\cmidrule(lr){1-2} \cmidrule(lr){3-4} \cmidrule(lr){5-6}
Total dialogues & 10,182 & Total utterances & 94,668 & Total tokens & 34,764,374 \\
\quad - Training & 6,000 & \multicolumn{2}{l}{Utterances per dialogue:} & \multicolumn{2}{l}{Tokens per utterance:} \\
\quad - Validation & 2,182 & \quad - Maximum & 30 & \quad - Maximum & 2,569 \\
\quad - Test & 2,000 & \quad - Average & 9.3 & \quad - Average & 367.22 \\
Multi-turn & 4,433 & \quad - Minimum & 2 & \quad - Minimum & 3 \\
Single-turn & 5,749 & & & & \\
\bottomrule[1.1pt]
\end{tabular}
\caption{Statistics of RETAIL at dialogue, utterance, and token levels.}
\label{tab:dataset_stats}
\end{table*}

For implicit queries commonly encountered in travel planning, we implement a decision-making support mechanism that mirrors natural human consultation processes. Using GPT-4o~\cite{openai2023}, we simulate tourist-agent interactions to reflect how users gradually articulate their travel needs and how agents guide the decision-making process. 

\textbf{Basic Information Clarification.} To systematically clarify basic travel requirements, the assistant combines systematic information gathering of essential planning elements (location, target cities, dates, and etc.) with natural conversation management strategies, limiting questions to 1-2 key points per round while employing colloquial language and context-adaptive dialogue progression.

\textbf{POI Clarification.} To address the diverse ways real tourists express their preferences, we implement comprehensive clarification logics considering multiple real-world factors. These logics simulate real-world travel planning scenarios by considering real-world factors like personal interests, weather conditions, POI's feedback. The assistant also provides reviews, ratings and high-quality images, enabling UGC-enriched decision-making through authentic dialogue interactions.

\subsubsection{Plan Generation and Revision}
To simulate the complexity and adaptability of real-world travel planning, our dataset encompasses both direct generation for explicit queries and dialogue-based planning for implicit queries. To capture the iterative nature of travel planning, we incorporate diverse plan revision cases spanning multiple aspects (dining, transportation, budget, weather-dependent activities). Detailed implementation specifications are elaborated in Section~\ref{method}.

\subsection{Dataset Annotations}

To ensure data integrity, we implement rigorous quality control measures across three key aspects: Tourism Knowledge Base Validation, Decision-Making Support Refinement, and Plan Generation and Revision Verification. Detailed information can be found in Appendix~\ref{appendix C}.

Our annotation team comprised 5 professionals (aged 25--40; 3 males, 2 females) with diverse expertise: two held master's degrees in linguistics, two held bachelor's degrees in computer science, and one specialized in psychology. All annotators received competitive compensation aligned with industry standards and their respective qualifications, ensuring fairness and motivation.

\textbf{Tourism Knowledge Base Validation.} To establish a robust knowledge base, we conduct comprehensive data cleaning by removing entries with missing key fields (POI names, locations, operating hours), implementing random replacement algorithms for sensitive information (contact details, user identifiers), and validating transportation data through manual field investigations of actual ticket prices. We supplement inter-city transportation data with optimized transfer routes where direct connections are unavailable, ensuring complete coverage of travel options.

\textbf{Decision-Making Support Refinement.} To ensure authentic travel planning interactions, we evaluate dialogues through multiple quality dimensions. We filter conversations with misaligned recommendations, ignored user needs, redundant suggestions, unnatural transitions, mechanical reasoning, inconsistent styles, and insufficient clarifications. Through prompt optimization and dialogue flow enhancement, we refine the dialogue to better mirror natural conversation patterns.

\textbf{Plan Generation and Revision Verification.} To ensure feasibility of travel plans, we filter plans with unreasonable schedules , impractical accommodations, unrealistic transportation, and conflicting activities. Plan revisions balance attraction and transportation adjustments while maintaining feasibility and user preferences, ensuring diverse practical travel plans.

\textbf{Dataset Quality.} Following \citet{shi2023midmed} and \citet{liu2022go}, we conduct human evaluation on 200 randomly sampled dialogues, where RETAIL achieves an average quality score of 0.85, demonstrating its high quality.

\subsection{Dataset Analysis}
Table~\ref{tab:dataset_stats} presents the statistics of RETAIL, which contains 10,182 dialogues with a balanced distribution between single-turn (5,749) and multi-turn (4,433) interactions, partitioned into training (6,000), validation (2,182), and test (2,000) sets. The dataset comprises 94,668 utterances and over 34 million tokens in total, representing substantial coverage of travel planning scenarios. Detailed statistics and examples of RETAIL can be found in Appendix~\ref{appendix C}.

\section{TGMA}
\label{method}

\begin{figure*}[t]
\centering
\includegraphics[width=\textwidth]{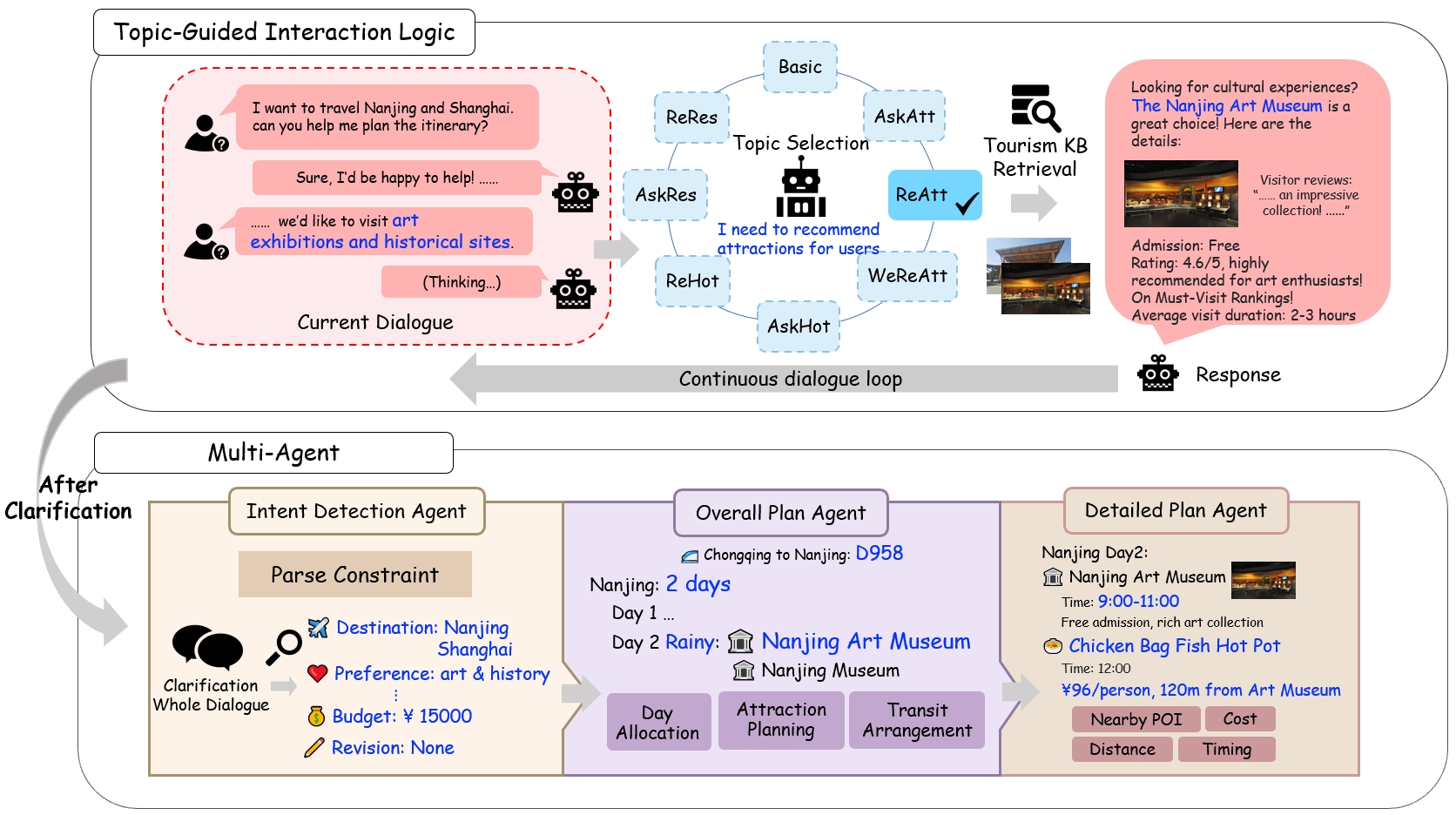}
\caption{Overview of TGMA framework: the Topic-Guided Interaction Logic handles user queries through topic selection and knowledge retrieval (top), while the Multi-Agent progresses from parsing user requirements to generating high-level plans (including transportation, durations, and attractions), ultimately producing an all-in-one travel plan with specific timing, locations, and other rich details (bottom).}
\end{figure*}

To mitigate the challenges of real-world travel planning, we propose TGMA framework, which combines Topic-Guided interaction with Multi-Agent planning. The Topic-Guided Interaction Logic selects appropriate topics based on dialogue context ensuring natural dialogue transitions aligned with user preferences, while the Multi-Agent architecture decomposes complex planning tasks into manageable steps for quality assurance.

\subsection{Topic-Guided Interaction Logic}
To address the challenge of maintaining coherent multi-turn dialogues in travel consultation, we propose a Topic-Guided Interaction Logic framework. It implements a structured dialogue management approach through an LLM-based topic selection mechanism, where eight interconnected states dynamically guide conversation flow based on dialogue context and user preferences. Our framework integrates dialogue context tracking with a comprehensive tourism knowledge base covering five essential domains (Attraction, Restaurant, Hotel, Transportation, and Weather). This design enables context-aware response generation while maintaining dialogue coherence, allowing the system to adapt to diverse user requirements through systematic topic transitions and knowledge integration.

\subsection{Multi-Agent}
To decompose complex travel planning into manageable specialized tasks, we design a multi-agent architecture that orchestrates three agents. This architecture enables progressive refinement of travel plans, from initial requirement analysis to detailed execution planning.

\textbf{The Intent Detection Agent} serves as the foundation by transforming unstructured conversational dialogues into structured representations. Through semantic parsing, it extracts and normalizes key travel parameters including location/time constraints, traveler details, accommodation preferences, and budget specifications, while also capturing users' modifications and refinements throughout the conversation. This maintains dialogue context and provides standardized inputs for subsequent planning stages.

\textbf{The Overall Plan Agent} leverages the ReAct\cite{yao2022react} framework to coordinate high-level planning decisions. It systematically processes Day Allocation, Attraction Planning, and Transit Arrangement through an integrated toolset for plan generation, analysis, and information retrieval. This structured approach ensures that generated plans effectively balance planning objectives with operational constraints, delivering coherent and executable travel recommendations.

\textbf{The Detailed Plan Agent} finalizes the planning process by transforming high-level outlines into comprehensive, executable plans. It systematically enriches travel outlines with nearby POI information and practical details, including attraction tickets, accommodation options, and restaurant recommendations. Through structured scheduling and point-of-interest coordination, the agent generates optimized daily arrangements that balance travel efficiency with user comfort while considering practical limitations.

\section{Experiments}

\begin{table*}[!htb]
\centering
\setlength{\tabcolsep}{8pt}
\renewcommand{\arraystretch}{0.9}
\begin{tabular}{l|cccc|c|c|c}
\toprule
\multirow{2}{*}{Model} & \multicolumn{4}{c|}{BLEU} & \multirow{2}{*}{METEOR} & \multirow{2}{*}{ROUGE-L} & \multirow{2}{*}{CIDEr} \\
\cmidrule{2-5}
& B-1 & B-2 & B-3 & B-4 & & & \\
\midrule

\rowcolor{blue!5} \multicolumn{8}{c}{\textbf{\textit{Baselines}}} \\
\midrule
\rowcolor{gray!5} Qwen2.5-7B-Instruct & 8.29 & 3.23 & 1.44 & 0.79 & 17.26 & 15.34 & 0.343 \\
\rowcolor{gray!5} Meta-Llama-3-8B-Instruct & 1.24 & 0.45 & 0.21 & 0.13 & 11.21 & 13.54 & 0.329 \\
\rowcolor{gray!5} glm-4-9b-chat & 7.54 & 3.90 & 1.94 & 1.18 & 17.09 & 15.14 & 0.348 \\
\rowcolor{gray!5} Baichuan2-13B-Chat & 9.49 & 4.41 & 1.76 & 0.96 & 14.74 & 15.49 & 0.345 \\
\rowcolor{gray!5} DeepSeek-V3(671B) & 1.13 & 0.69 & 0.36 & 0.24 & 13.19 & 13.69 & 0.333 \\
\rowcolor{yellow!5} Doubao-Pro & 1.77 & 0.26 & 0.10 & 0.00 & 11.58 & 13.60 & 0.328 \\
\rowcolor{yellow!5} Baichuan3-Turbo & 2.81 & 1.44 & 0.79 & 0.52 & 16.43 & 14.21 & 0.337 \\
\rowcolor{yellow!5} Qwen-Plus & 7.33 & 3.99 & 2.08 & 1.32 & 16.81 & 15.07 & 0.346 \\
\rowcolor{yellow!5} gpt-4-Turbo & 4.91 & 2.67 & 1.52 & 1.03 & 14.18 & 15.10 & 0.347 \\
\rowcolor{yellow!5} gpt-4o & 7.39 & 3.34 & 1.63 & 0.99 & 15.40 & 15.31 & 0.342 \\
\midrule
\rowcolor{blue!5} \multicolumn{8}{c}{\textbf{\textit{Topic-Guided Interaction Logic(ours)}}} \\
\midrule
\rowcolor{gray!5} DeepSeek-V3(671B) & 13.43 & \textbf{6.86} & \textbf{4.16} & \textbf{2.91} & 18.40 & 15.98 & 0.354 \\
\rowcolor{gray!5} Qwen2.5-7B-Instruct(SFT) & \textbf{13.88} & 6.49 & 3.32 & 2.04 & 21.44 & 16.33 & \textbf{0.360} \\
\rowcolor{yellow!5} gpt-4o & 10.96 & 4.50 & 2.29 & 1.36 & \textbf{22.05} & \textbf{16.85} & 0.343 \\
\bottomrule
\end{tabular}
\caption{Decision-Making Support Evaluation. ROUGE, BLEU, and METEOR metrics are expressed as percentages (\%). Gray and yellow backgrounds denote open-source and closed-source models.}
\label{table:dialog_comparison}
\end{table*}

\begin{table*}[!htb]
\normalsize
\setlength{\tabcolsep}{6pt}
\renewcommand{\arraystretch}{0.9}
\centering
\begin{tabular}{l|cccc|cccc|c}
\toprule
\multirow{4}{*}{Model} & \multicolumn{4}{c|}{Commonsense Constraint} & \multicolumn{4}{c|}{User Preference Constraint} & \multirow{4}{*}{\makecell{Final Pass\\Rate}} \\
\cmidrule(lr){2-5} \cmidrule(lr){6-9}
& \multicolumn{2}{c}{Micro} & \multicolumn{2}{c|}{Macro} & \multicolumn{2}{c}{Micro} & \multicolumn{2}{c|}{Macro} & \\
\cmidrule(lr){2-3} \cmidrule(lr){4-5} \cmidrule(lr){6-7} \cmidrule(lr){8-9}
& PR. & F1 & PR. & F1 & PR. & F1 & PR. & F1 & \\
\midrule

\rowcolor{blue!5} \multicolumn{10}{c}{\textbf{\textit{Baselines}}} \\
\midrule
\rowcolor{gray!5} Qwen2.5-7B-Instruct & 49.38 & 0.50 & 0.00 & 0.32 & 74.49 & 0.74 & 22.87 & 0.42 & 0.00 \\
\rowcolor{gray!5} Meta-Llama-3-8B-Instruct & 64.60 & 0.65 & 0.00 & 0.38 & 68.04 & 0.68 & 17.42 & 0.40 & 0.00 \\
\rowcolor{gray!5} glm-4-9b-chat & 54.13 & 0.54 & 0.17 & 0.35 & 77.32 & 0.77 & 29.69 & 0.43 & 0.10 \\
\rowcolor{gray!5} Baichuan2-13B-Chat & 50.00 & 0.50 & 0.00 & 0.33 & 60.42 & 0.75 & 0.00 & 0.43 & 0.00 \\
\rowcolor{gray!5} DeepSeek-V3(671B) & 70.55 & 0.74 & 2.92 & 0.42 & 79.73 & 0.80 & 35.27 & 0.44 & 0.81 \\[0.5ex]
\rowcolor{yellow!5} Doubao-Pro & 62.30 & 0.62 & 0.00 & 0.38 & 75.94 & 0.76 & 26.32 & 0.43 & 0.00 \\
\rowcolor{yellow!5} Baichuan3-Turbo & 70.72 & 0.75 & 0.08 & 0.39 & 65.41 & 0.65 & 9.62 & 0.39 & 0.00 \\
\rowcolor{yellow!5} Qwen-Plus & 75.17 & 0.79 & 3.84 & 0.44 & 79.95 & 0.79 & 35.91 & 0.44 & 1.00 \\
\rowcolor{yellow!5} gpt-4-Turbo & 68.43 & 0.64 & 0.35 & 0.39 & 77.45 & 0.79 & 30.00 & 0.44 & 0.20 \\
\rowcolor{yellow!5} gpt-4o & 71.84 & 0.73 & 0.36 & 0.42 & 78.55 & 0.79 & 32.90 & 0.44 & 0.10 \\
\midrule
\rowcolor{blue!5} \multicolumn{10}{c}{\textbf{\textit{TGMA(ours)}}} \\
\midrule
\rowcolor{gray!5} DeepSeek-V3(671B) & 79.64 & 0.78 & 5.73 & 0.43 & 82.09 & 0.84 & 39.49 & 0.45 & 1.38 \\
\rowcolor{gray!5} Qwen2.5-7B-Instruct(SFT) & \textbf{81.74} & 0.77 & \textbf{5.89} & \textbf{0.45} & \textbf{86.40} & \textbf{0.84} & \textbf{42.32} & \textbf{0.46} & \textbf{2.72} \\
\rowcolor{yellow!5} gpt-4o & 78.65 & \textbf{0.79} & 3.49 & 0.44 & 80.30 & 0.81 & 37.30 & 0.45 & 1.07 \\
\bottomrule
\end{tabular}
\caption{Travel Plan Evaluation. Pass Rate (PR.) metrics are reported in percentages (\%). Gray and yellow backgrounds denote open-source and closed-source models.}
\label{tab:plan_comparison_1}
\end{table*}

\subsection{Baselines}

Following \citet{xie2024travelplanner} and \citet{shao2024chinatravel}, we carefully select a few strong baselines for comparison, including GPT-4o~\cite{openai2023}, GPT-4-Turbo~\cite{achiam2023gpt}, DeepSeek-V3~\cite{liu2024deepseek}, and Llama-3~\cite{room2024llama}, as well as Chinese models like Baichuan~\cite{lin2024baichuan}, GLM~\cite{glm2024chatglm}, and Qwen~\cite{bai2023qwen}.

\begin{figure*}[t]
\centering
\includegraphics[width=0.95\textwidth]{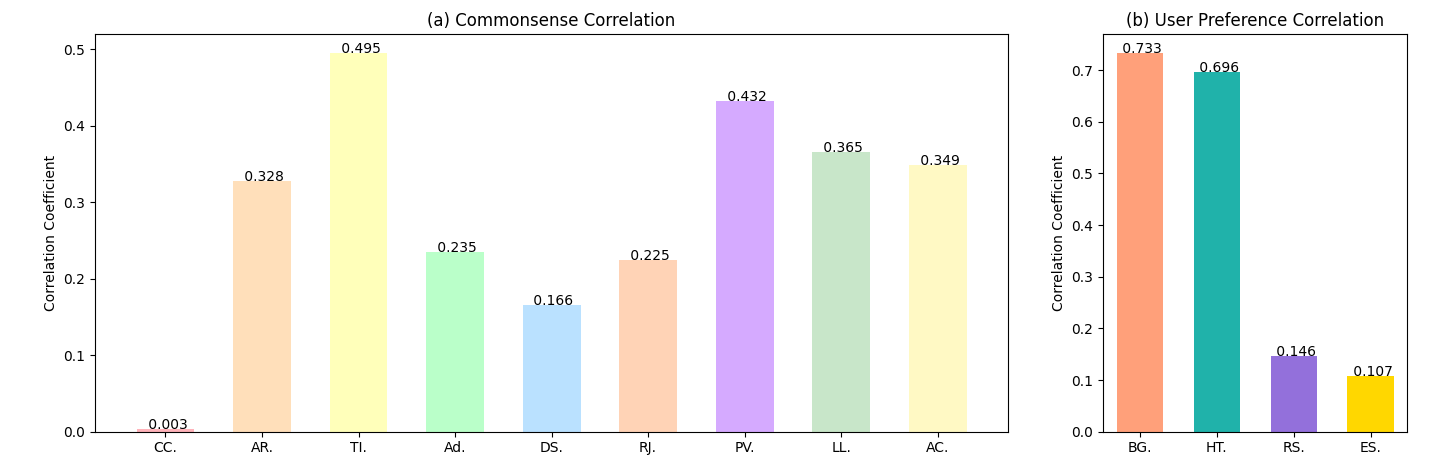}
\caption{Impact of Different Constraints on Pass Rates. (a) CC: City Coverage, AR: Activity Repetition, TI: Time Interval, Ad: Accommodation, DS: Daily Schedule, RJ: Return Journey, PV: POI Validation, LL: Location Logic, AC: Activity Count. (b) Bg: Budget, HT: Hotel Type, RS: Required Sites, ES: Excluded Sites.}
\label{fig:constraints_impact}
\end{figure*}

\subsection{Metrics}
We assess the system from two aspects: decision-making process's dialogue capabilities and planning performance. \textbf{For dialogue capabilities,} following \citet{shi2023midmed} and \citet{liu2022go}, we employ standard dialogue quality metrics including BLEU-n~\cite{papineni2002bleu}, ROUGE~\cite{lin2004rouge} and METEOR~\cite{banerjee2004meteor} to assess the response generation. \textbf{For planning performance}, following ~\citet{xie2024travelplanner}, we evaluate plans using commonsense constraint and user preference constraint pass rate, and final pass rate. Detailed definition can be found in Appendix ~\ref{appendix A.1}. 

\subsection{Decision-Making Support Evaluation}

The evaluation results in Table~\ref{table:dialog_comparison} demonstrate the effectiveness of our Topic-Guided Interaction Logic framework. When integrated with different LLMs, our framework consistently outperforms their baseline across all metrics. Specifically, with DeepSeek-V3, our framework achieves substantial improvements in BLEU scores, while with GPT-4, it shows significant gains in METEOR (22.05 vs 15.40) and ROUGE-L (16.85 vs 15.31). These improvements indicate that our structured dialogue management approach effectively enhances model performance in travel-specific conversations. Detailed experimental information can be found in Appendix~\ref{appendix A.3}.

\subsection{Travel Plan Evaluation}
\label{sec:travel_eval}

Table~\ref{tab:plan_comparison_1} demonstrates the superior real-world planning capabilities of TGMA. When integrated with Qwen2.5-7B-Instruct(SFT), our approach achieves the highest final pass rate of 2.72\%, significantly outperforming baseline models across both commonsense and user preference constraints. The notable performance gap between our framework and baseline models highlights the effectiveness of our structured planning approach in handling real-world travel planning scenarios. Detailed experimental information can be found in Appendix~\ref{appendix A.4}.

\section{Analysis}
\subsection{RQ1: How Do Different Constraints Impact Travel Plan Final Pass Rate?}
To study \textbf{how different constraints affect the final pass rate of travel plans} which is essential for system optimization, we use the travel plans to calculate the Pearson correlation coefficients between 14 types of constraints (9 commonsense constraints and 4 user preference constraints) shown in Table~\ref{tab:constraints}.

Figure \ref{fig:constraints_impact}(a) shows Time Interval and POI Validation are the main constraints affecting pass rates, at 0.495 and 0.432 respectively. This suggests plans need better activity scheduling and POI verification. City Coverage at 0.003 has minimal impact, indicating most plans satisfy city coverage requirements.

Figure \ref{fig:constraints_impact}(b) reveals Hotel Type and Required Sites as key preference constraints at 0.733 and 0.696, while Budget and Excluded Sites show lower impact at 0.146 and 0.107. This indicates plans mainly struggle with meeting hotel type and must-visit site requirements.

\section{Conclusion}
This work identifies three critical limitations in real-world travel planning: insufficient decision-making support for implicit queries, the lack of environmental awareness, and the inability to generate all-in-one plans with rich details. To mitigate these challenges, we first introduce RETAIL covering both explicit and decision-making support for implicit queries. It also features environmental awareness, and rich POI information enabling all-in-one travel plans. Based on RETAIL, we propose TGMA, a topic-guided multi-agent framework for providing coherent and user-aligned decision support and generating all-in-one travel plans that integrate various practical aspects. Through extensive experiments, we demonstrate TGMA's effectiveness in real-world travel planning scenarios. We hope our work provides valuable insights and resources for advancing real-world travel planning systems.

\section*{Limitations}
RETAIL requires substantial computational resources and poses significant challenges to models' context processing capabilities, especially when handling extensive text content and maintaining contextual understanding throughout long-form interactions. Future work will explore Progressive Generation\cite{P-RAG} to mitigate these computational constraints and enhance context retention in extended sequences.

\section*{Ethics Statement}
We make sure that RETAIL is collected in a manner that is consistent with the terms of use of any sources and the intellectual property and privacy rights of the original authors of the texts. All participants provided informed consent and were fairly compensated. Details of data collection procedures are described in Section~\ref{sec:dataset_construct}.

\bibliography{main}
\bibliographystyle{acl_natbib}

\appendix

\appendix

\section*{Appendix}

\begin{table*}[tb]
\centering
\renewcommand{\arraystretch}{1.3}
\setlength{\tabcolsep}{8pt}
\begin{tabular}{lll}
\toprule
\textbf{Category} & \textbf{Metric} & \textbf{Description} \\
\midrule
\multirow{10}{*}{\makecell[l]{\textbf{Commonsense}\\\textbf{Constraints}}} 
& City Coverage & Verifies all target cities are included \\
& Activity Repetition & \textcolor{brown}{Detects duplicate activities in itinerary} \\
& Time Interval & \textcolor{brown}{Verifies 30-minute minimum gaps between activities} \\
& Accommodation & \textcolor{brown}{Ensures lodging is arranged for all nights except last} \\
& Daily Schedule & Validates completion of daily activity planning \\
& Return Journey & Confirms return trip on final day \\
& POI Validation & \textcolor{brown}{Verifies existence of all points of interest} \\
& Location Logic & \textcolor{brown}{Ensures POIs are within designated cities} \\
& Activity Count & \textcolor{brown}{Verifies minimum 4 activities per city daily} \\  
& & \textcolor{brown}{(\textbf{including three meals a day, attractions, and snack shops})} \\ 
\midrule
\multirow{4}{*}{\makecell[l]{\textbf{User}\\\textbf{Preferences}}} 
& Budget & Validates total cost against user budget \\
& Hotel Type & \textcolor{brown}{Confirms compliance with hotel preferences} \\
& Required Sites & \textcolor{brown}{Verifies inclusion of must-visit attractions} \\
& Excluded Sites & \textcolor{brown}{Ensures exclusion of unwanted locations} \\
\bottomrule
\end{tabular}
\caption{Evaluation metrics for Travel Plan Evaluation. Metrics in black are common to both TravelPlanner~\cite{xie2024travelplanner} and our benchmark, while those in \textcolor{brown}{brown} are unique to our benchmark.}
\label{tab:constraints}
\end{table*}

\section{Additional Experiment}

\subsection{Detailed Evaluation Metrics}
\label{appendix A.1}

To ensure the quality and practicality of generated travel itineraries, we establish a comprehensive evaluation framework comprising two main categories of metrics, as detailed in Table~\ref{tab:constraints}. The first category, Commonsense Constraints, consists of ten fundamental metrics that verify the logical coherence and feasibility of travel plans. These metrics range from basic requirements such as city coverage and activity repetition checks, to more sophisticated validations like temporal conflict detection and location logic verification. The second category focuses on User Preferences, incorporating four key metrics to assess how well the generated itineraries align with specific user requirements, including budget constraints and accommodation preferences. Notably, our benchmark introduces several novel metrics (marked in brown) that were not present in previous evaluation frameworks, particularly enhancing the assessment of spatial-temporal consistency and user preference compliance. This expanded set of metrics enables a more rigorous and comprehensive evaluation of travel planning systems.

\subsection{Additional Analysis}
\label{appendix A.2}

\subsubsection{RQ2: How Important is Decision-Making Support in Travel Planning?}
To evaluate \textbf{the importance of decision-making support}, we employed a dual-assessment approach combining \textbf{automated metrics} and \textbf{human evaluations} to compare travel plans generated with and without assisted decision-making processes.

As for \textbf{automated metrics}, we employed the same evaluation metrics used in Travel Plan Evaluation~\ref{sec:travel_eval} to assess plans generated with and without assisted decision-making processes. The results in Table~\ref{tab:1} demonstrate significant performance gaps across all metrics. Plans without decision-making support achieves only half the micro and macro pass rates for commonsense constraints compared to those with support. The gap widens further for user preference constraints, where unsupported plans showed substantially lower performance in both micro and macro evaluations. Without decision-making support, models tend to generate generic travel plans that fail to capture individual preferences and requirements, resulting in standardized itineraries that lack personalization. This limitation lead to zero final passes, highlighting that decision-making support is crucial for understanding user preferences and creating personalized travel experiences.
\begin{table}[!htb]
\small
\setlength{\tabcolsep}{6pt}
\renewcommand{\arraystretch}{0.9}
\centering
\begin{tabular}{l|cc|cc|c}
\toprule
\multirow{3}{*}{DMS} & \multicolumn{2}{c|}{CC} & \multicolumn{2}{c|}{UPC} & \multirow{3}{*}{\makecell{Final\\PR (\%)}} \\
\cmidrule(lr){2-3} \cmidrule(lr){4-5}
& \multicolumn{1}{c}{Micro} & \multicolumn{1}{c|}{Macro} & \multicolumn{1}{c}{Micro} & \multicolumn{1}{c|}{Macro} & \\
\midrule
\xmark & 34.19 & 2.60 & 31.49 & 1.17 & 0.00 \\
\cmark & 78.65 & 3.49 & 80.30 & 37.30  & 1.07 \\
\bottomrule
\end{tabular}
\caption{Travel plan evaluation on CC (Commonsense Constraints) and UPC (User Preference Constraints). DMS: Decision-Making Support; PR: Pass Rate.}
\label{tab:1}
\end{table}

As for \textbf{human evaluations}, we employed the same team of 5 expert annotators (previously used for dataset annotation). The annotators demonstrated strong agreement during dataset annotation, with a Cohen’s kappa coefficient exceeding 0.9, ensuring high consistency in their evaluations.

We designed three 5-point metrics for evaluation. 
\begin{itemize}
\item \textbf{Implicit Need Alignment} measures the system's ability to address unstated requirements, where 5 points indicate full coverage of contextual needs through proactive suggestions, and 1 point reflects complete oversight of implicit preferences.

\item \textbf{Revision Cost} quantifies plan refinement effort, where 5 points represent immediate acceptance (0--1 edits required), and 1 point indicates extensive rework (5+ edits needed) to meet basic requirements.

\item \textbf{Overall Satisfaction} captures holistic user experience, where 5 points reflect perfect alignment with both stated and unstated needs, while 1 point indicates fundamental dissatisfaction with the proposed itinerary.
\end{itemize}

Table~\ref{tab:dms_comparison} showed statistically significant improvements across all metrics with decision-making support. Implicit need alignment rose from 2.1 to 3.9, with experts noting better recognition of contextual constraints (e.g., mobility requirements). Baseline plans required 2.1 average edits versus 1.2 with support, though three users still needed multiple adjustments for niche preferences. Four participants described the clarification process as "useful for refining initial ideas," while one found it excessive for simple day trips. Time savings averaged 28 minutes per task, with greater benefits for multi-city itineraries.

\begin{table}[!htb]
\small
\setlength{\tabcolsep}{6pt}
\renewcommand{\arraystretch}{0.9}
\centering
\begin{tabular}{lccc}
\toprule
\textbf{Setting} & \textbf{Need Align.} & \textbf{Rev. Cost} & \textbf{Satisfaction} \\
\midrule
w/ DMS & 3.9 & 1.2 & 4.2 \\
w/o DMS & 2.1 & 2.1 & 3.1 \\
\bottomrule
\end{tabular}
\caption{Performance comparison with/without Decision-Making Support (DMS). Need Align.: Implicit Need Alignment; Rev. Cost: Revision Cost; Satisfaction: Overall Satisfaction}
\label{tab:dms_comparison}
\end{table}

\subsubsection{RQ3: How Do Multimodal Information and User-Generated Content Enhance Travel Planning?}
To evaluate \textbf{the impact of multimodal information and user-generated content on travel planning quality}, we employed the same team of 5 expert annotators (previously used for dataset annotation) to assess four system settings: basic text-only setting, enhanced with user reviews, enhanced with visual content, and enhanced with both elements. The annotators demonstrated strong agreement during dataset annotation, with a Cohen’s kappa coefficient exceeding 0.9, ensuring high consistency in their evaluations.

We designed four evaluation metrics on a 5-point scale. 
\begin{itemize}
\item \textbf{Information Comprehension} measures users' understanding of travel destinations, where 5 points indicate complete and clear understanding of destination features, and 1 point reflects confusion or misunderstanding.

\item \textbf{Decision Efficiency} evaluates the speed and quality of travel decisions, where 5 points represent quick and confident decision-making, and 1 point indicates hesitation and uncertainty.

\item \textbf{Interaction Experience} assesses the effectiveness of plan modification process, where 5 points reflect smooth and intuitive plan adjustments with clear feedback, and 1 point suggests difficult and confusing modification experience.

\item \textbf{Overall Satisfaction} captures the comprehensive user experience, where 5 points indicate highly satisfactory planning experience, and 1 point reflects poor user satisfaction.
\end{itemize}

The experimental results demonstrate that both visual content and user-generated content significantly enhance the travel planning process. While the basic text-only setting achieves moderate performance, the addition of either visual content or user reviews substantially improves user experience across all metrics. The combination of both elements yields the best performance, achieving scores above 4.0 across all dimensions, which validates the importance of incorporating multimodal information in travel planning systems.

\begin{table}[h]
\small
\centering
\setlength{\tabcolsep}{4pt}
\renewcommand{\arraystretch}{1.2}
\begin{tabular}{l cccc}
\toprule
Metrics & Basic & +UGC & +Visual & +Both \\
\midrule
Info. Comprehension & 3.28 & 3.85 & 3.88 & 4.12 \\
Decision Efficiency & 3.45 & 3.92 & 3.95 & 4.15 \\
Interaction Exp. & 3.52 & 3.89 & 3.87 & 4.08 \\
Overall Satisfaction & 3.31 & 3.88 & 3.85 & 4.16 \\
\bottomrule
\end{tabular}
\caption{Impact of multimodal information and user-generated content on travel planning quality. Basic: text-only setting; +UGC: enhanced with user reviews; +Visual: enhanced with visual content; +Both: enhanced with both elements. All scores range from 0-5, averaged over user ratings.}
\label{tab:user_study}
\end{table}

\subsubsection{RQ4: To What Extent Do Environmental Factors Influence Travel Planing?}

To evaluate \textbf{the impact of environmental factors on travel planning}, we employed the same team of 5 expert annotators (previously used for dataset annotation) to assess travel plans generated with and without environmental awareness. Each annotator evaluated 20 different scenarios, resulting in 100 test cases in total. The annotators demonstrated strong agreement during dataset annotation, with a Cohen’s kappa coefficient exceeding 0.9, ensuring high consistency in their evaluations.

We designed three 5-point metrics for evaluation. 
\begin{itemize}
\item \textbf{Weather Adaptability Score} measures plan adaptation to weather conditions, where 5 points indicate comprehensive weather-based adjustments and alternatives, while 1 point reflects minimal weather consideration.

\item \textbf{Transportation Accessibility Score} evaluates traffic-aware planning, where 5 points represent detailed routing with backup options, and 1 point indicates limited traffic consideration.

\item \textbf{Environmental Alert Score in Requirement Clarification} assesses proactive environmental guidance, where 5 points represent timely alerts with practical suggestions, and 1 point reflects delayed or missing environmental alerts.
\end{itemize}

The experimental results demonstrate that environmental awareness significantly enhances travel planning quality. With environmental awareness enabled, all three metrics show notable improvements, particularly in the Environmental Alert Score (from 3.47 to 4.20). These results validate the importance of incorporating environmental factors in travel planning systems.

\begin{table}[!htb]
\setlength{\tabcolsep}{9pt}
\renewcommand{\arraystretch}{1.05}
\centering
\begin{tabular}{l|ccc}
\toprule
EA & WAS & TAS & EARC \\
\midrule
\xmark & 3.85 & 3.72 & 3.47 \\
\cmark & 4.20 & 4.00 & 4.20 \\
\bottomrule
\end{tabular}
\caption{Impact of Environmental Awareness (EA) on travel planning quality. WAS: Weather Adaptability Score; TAS: Transportation Accessibility Score; EARC: Environmental Alert Score in Requirement Clarification. All scores range from 0-5, averaged over 100 test cases.}
\label{tab:env_aware}
\end{table}

\subsection{Detailed Decision-Making Support Evaluation}
\label{appendix A.3}
\begin{table*}[tp]
\centering
\setlength{\tabcolsep}{9pt}
\renewcommand{\arraystretch}{1.3}
\begin{tabular}{l|cccc|ccc}
\toprule
\multirow{2}{*}{Model} & \multicolumn{4}{c|}{BLEU} & \multirow{2}{*}{METEOR} & \multirow{2}{*}{ROUGE-L} & \multirow{2}{*}{CIDEr} \\
\cmidrule{2-5}
& B-1 & B-2 & B-3 & B-4 & & & \\
\midrule
\rowcolor{blue!5} \multicolumn{8}{c}{\textbf{One-shot In-Context Learning}} \\
\midrule
\rowcolor{blue!5} \multicolumn{8}{l}{\textbf{\textit{Baselines}}} \\
\rowcolor{gray!5} Qwen2.5-7B-Instruct & 9.67 & 2.78 & 1.10 & 0.61 & 19.91 & 17.89 & 0.353 \\
\rowcolor{gray!5} Meta-Llama-3-8B-Instruct & 2.10 & 0.50 & 0.15 & 0.06 & 10.62 & 13.53 & 0.328 \\
\rowcolor{gray!5} glm-4-9b-chat & 1.11 & 0.57 & 0.21 & 0.10 & 13.54 & 13.84 & 0.333 \\
\rowcolor{gray!5} Baichuan2-13B-Chat & 8.82 & 2.63 & 0.96 & 0.53 & 15.47 & 16.51 & 0.350 \\
\rowcolor{gray!5} DeepSeek-V3(671B) & 0.35 & 0.21 & 0.13 & 0.10 & 13.28 & 14.45 & 0.332 \\

\rowcolor{yellow!5} Doubao-Pro & 6.06 & 1.00 & 0.38 & 0.20 & 12.22 & 15.04 & 0.335 \\
\rowcolor{yellow!5} Baichuan3-Turbo & 2.41 & 1.15 & 0.58 & 0.40 & 15.91 & 14.14 & 0.337 \\
\rowcolor{yellow!5} Qwen-Plus & 13.37 & 5.60 & 2.37 & 1.35 & 19.25 & 16.38 & 0.350 \\
\rowcolor{yellow!5} gpt-4-Turbo & 5.30 & 2.31 & 1.25 & 0.85 & 15.00 & 15.22 & 0.341 \\
\rowcolor{yellow!5} gpt-4o & 9.39 & 3.30 & 1.31 & 0.69 & 15.22 & 15.47 & 0.339 \\
\rowcolor{blue!5} \multicolumn{8}{l}{\textbf{\textit{Topic-Guided Interaction Logic(ours)}}} \\
\rowcolor{gray!5}DeepSeek-V3(671B) & 5.96 & 2.61 & 1.16 & 0.65 & 16.83 & 14.95 & 0.342 \\
\rowcolor{gray!5}Qwen2.5-7B-Instruct(SFT) & \textbf{14.20} & \textbf{6.70} & \textbf{3.45} & \textbf{2.15} & 21.80 & 16.50 & \textbf{0.362} \\
\rowcolor{yellow!5} gpt-4o & 12.37 & 4.86 & 1.98 & 0.89 & \textbf{22.43} & \textbf{16.95} & 0.349 \\

\midrule[\heavyrulewidth]

\rowcolor{blue!5} \multicolumn{8}{c}{\textbf{Two-shot In-Context Learning}} \\
\midrule
\rowcolor{blue!5} \multicolumn{8}{l}{\textbf{\textit{Baselines}}} \\
\rowcolor{gray!5}Qwen2.5-7B-Instruct & 7.75 & 2.21 & 0.77 & 0.37 & 19.56 & 18.17 & 0.353 \\
\rowcolor{gray!5}Meta-Llama-3-8B-Instruct & 1.37 & 0.31 & 0.10 & 0.05 & 10.34 & 13.48 & 0.328 \\
\rowcolor{gray!5}glm-4-9b-chat & 0.70 & 0.36 & 0.13 & 0.07 & 12.96 & 13.62 & 0.331 \\
\rowcolor{gray!5}Baichuan2-13B-Chat & 10.16 & 3.59 & 1.33 & 0.69 & 15.20 & 15.99 & 0.349 \\
\rowcolor{gray!5}DeepSeek-V3(671B) & 0.29 & 0.17 & 0.12 & 0.09 & 13.30 & 13.47 & 0.336 \\

\rowcolor{yellow!5}Doubao-Pro & 2.89 & 0.81 & 0.34 & 0.17 & 14.01 & 14.15 & 0.334 \\
\rowcolor{yellow!5}Baichuan3-Turbo & 1.95 & 0.88 & 0.45 & 0.30 & 15.85 & 14.07 & 0.336 \\
\rowcolor{yellow!5}Qwen-Plus & 12.30 & 5.14 & 2.04 & 1.12 & 19.16 & 16.14 & 0.347 \\
\rowcolor{yellow!5}gpt-4-Turbo & 2.47 & 1.08 & 0.55 & 0.35 & 14.14 & 14.53 & 0.337 \\
\rowcolor{yellow!5}gpt-4o & 9.97 & 3.99 & 1.46 & 0.76 & 17.07 & 15.81 & 0.343 \\

\rowcolor{blue!5} \multicolumn{8}{l}{\textbf{\textit{Topic-Guided Interaction Logic(ours)}}} \\

\rowcolor{gray!5}DeepSeek-V3(671B) & 7.31 & 3.19 & 1.67 & 1.09 & 17.30 & 15.30 & 0.347 \\
\rowcolor{gray!5}Qwen2.5-7B-Instruct(SFT) & \textbf{14.40} & \textbf{6.90} & \textbf{3.60} & \textbf{2.25} & 22.00 & 16.65 & \textbf{0.364} \\
\rowcolor{yellow!5}gpt-4o & 12.86 & 4.72 & 1.98 & 1.02 & \textbf{22.29} & \textbf{16.76} & 0.343 \\

\bottomrule
\end{tabular}
\caption{Decision-Making Support Evaluation under One-shot and Two-shot In-Context Learning. ROUGE, BLEU, and METEOR metrics are expressed as percentages (\%). Gray and yellow backgrounds denote open-source and closed-source models.}
\label{table:results}
\end{table*}

As shown in Table \ref{table:results}, our Topic-Guided Interaction Logic approach demonstrates significant improvements in models' ability to clarify travel requirements. Our enhanced Qwen2.5-7B-Instruct with SFT achieves a remarkable 252\% improvement in BLEU-4 scores from 0.61 to 2.15 in one-shot learning, while maintaining stable performance in two-shot scenarios at 2.25. This consistency contrasts sharply with baseline models' fluctuations, exemplified by gpt-4-Turbo's BLEU-4 dropping from 0.85 to 0.35 between one-shot and two-shot settings.

The effectiveness is further validated by improvements across all metrics, with METEOR scores reaching 22.43 for gpt-4o and ROUGE-L achieving 16.95, indicating enhanced semantic understanding. Our method shows strong scalability, enabling smaller models like Qwen2.5-7B-Instruct to outperform larger baselines with a BLEU-1 score of 14.20. The consistent CIDEr score improvements across all enhanced models confirm the overall quality enhancement of generated responses.

These comprehensive improvements suggest our approach effectively addresses the core challenges in travel requirement clarification. The method's success across both open-source and closed-source models, along with its stability in different shot settings, demonstrates its potential as a robust solution for enhancing travel dialogue systems' ability to process ambiguous user requests.

\subsection{Detailed Travel Plan Evaluation}
\label{appendix A.4}
\begin{table*}[tp]
\centering
\setlength{\tabcolsep}{6pt}
\renewcommand{\arraystretch}{1}
\begin{threeparttable}
\begin{tabular}{l|cccc|cccc|c}
\toprule
\multirow{4}{*}{Model} & \multicolumn{4}{c|}{Commonsense Constraint} & \multicolumn{4}{c|}{User Preference Constraint} & \multirow{4}{*}{\makecell{Final Pass\\Rate}} \\
\cmidrule(lr){2-5} \cmidrule(lr){6-9} 
& \multicolumn{2}{c}{Micro} & \multicolumn{2}{c|}{Macro} & \multicolumn{2}{c}{Micro} & \multicolumn{2}{c|}{Macro} & \\
\cmidrule(lr){2-3} \cmidrule(lr){4-5} \cmidrule(lr){6-7} \cmidrule(lr){8-9}
& PR. & F1 & PR. & F1 & PR. & F1 & PR. & F1 & \\
\midrule
\rowcolor{blue!5} \multicolumn{10}{c}{\textbf{CoT}} \\
\midrule
\rowcolor{blue!5} \multicolumn{10}{l}{\textbf{\textit{Baselines}}} \\
\rowcolor{gray!5}Qwen2.5-7B-Instruct & 47.17 & 0.49 & 0.00 & 0.32 & 76.33 & 0.75 & 26.24 & 0.43 & 0.00 \\
\rowcolor{gray!5}Meta-Llama-3-8B-Instruct & 63.03 & 0.63 & 0.00 & 0.39 & 72.41 & 0.72 & 24.55 & 0.42 & 0.00 \\
\rowcolor{gray!5}glm-4-9b-chat & 51.66 & 0.52 & 0.21 & 0.33 & 77.53 & 0.77 & 30.30 & 0.43 & 0.10 \\
\rowcolor{gray!5}Baichuan2-13B-Chat & 55.00 & 0.55 & 0.00 & 0.35 & 50.00 & 0.50 & 0.00 & 0.33 & 0.00 \\
\rowcolor{gray!5}DeepSeek-V3(671B) & 67.70 & 0.71 & 1.14 & 0.40 & 79.49 & 0.79 & 36.15 & 0.44 & 0.30 \\

\rowcolor{yellow!5}Doubao-Pro & 54.49 & 0.52 & 0.00 & 0.35 & 77.45 & 0.76 & 31.37 & 0.43 & 0.00 \\
\rowcolor{yellow!5}Baichuan3-Turbo & 70.54 & 0.71 & 0.00 & 0.41 & 65.94 & 0.66 & 10.77 & 0.39 & 0.00 \\
\rowcolor{yellow!5}Qwen-Plus & 68.34 & 0.70 & 0.65 & 0.41 & 80.18 & 0.80 & 37.42 & 0.44 & 0.23 \\
\rowcolor{yellow!5}gpt-4-Turbo & 74.46 & 0.70 & 2.14 & 0.42 & 77.72 & 0.77 & 32.03 & 0.43 & 0.87 \\
\rowcolor{yellow!5}gpt-4o & 77.17 & 0.77 & 3.51 & 0.43 & 78.99 & 0.79 & 33.77 & 0.44 & 0.90 \\

\rowcolor{blue!5} \multicolumn{10}{l}{\textbf{\textit{TGMA(ours)}}} \\
\rowcolor{gray!5} DeepSeek-V3(671B) & 79.81 & 0.78 & 5.76 & 0.43 & 80.02 & \textbf{0.87} & 42.24 & 0.45 & 1.41 \\
\rowcolor{gray!5} Qwen2.5-7B-Instruct(SFT) & \textbf{83.39} & 0.79 & \textbf{8.12} & \textbf{0.47} & \textbf{87.94} & 0.86 & \textbf{43.09} & \textbf{0.46} & \textbf{2.99} \\
\rowcolor{yellow!5} gpt-4o & 77.78 & \textbf{0.81} & 4.32 & 0.45 & 84.43 & 0.82 & 42.67 & 0.44 & 2.03 \\
\bottomrule
\end{tabular}
\caption{Travel Plan Evaluation under Chain-of-Thought(CoT) Setting. Pass Rate (PR.) metrics are reported in percentages (\%). Gray and yellow
backgrounds denote open-source and closed-source models.}
\label{tab:additional_plan_comparison_cot}
\end{threeparttable}
\end{table*}

\begin{table*}[tp]
\centering
\setlength{\tabcolsep}{6pt}
\renewcommand{\arraystretch}{1}
\begin{threeparttable}
\begin{tabular}{l|cccc|cccc|c}
\toprule
\multirow{4}{*}{Model} & \multicolumn{4}{c|}{Commonsense Constraint} & \multicolumn{4}{c|}{User Preference Constraint} & \multirow{4}{*}{\makecell{Final Pass\\Rate}} \\
\cmidrule(lr){2-5} \cmidrule(lr){6-9} 
& \multicolumn{2}{c}{Micro} & \multicolumn{2}{c|}{Macro} & \multicolumn{2}{c}{Micro} & \multicolumn{2}{c|}{Macro} & \\
\cmidrule(lr){2-3} \cmidrule(lr){4-5} \cmidrule(lr){6-7} \cmidrule(lr){8-9}
& PR. & F1 & PR. & F1 & PR. & F1 & PR. & F1 & \\
\midrule
\rowcolor{blue!5} \multicolumn{10}{c}{\textbf{One-shot In-Context Learning}} \\
\midrule

\rowcolor{blue!5} \multicolumn{10}{l}{\textbf{\textit{Baselines}}} \\
\rowcolor{gray!5}Qwen2.5-7B-Instruct & 58.26 & 0.58 & 0.08 & 0.37 & 79.92 & 0.79 & 29.95 & 0.44 & 0.08 \\
\rowcolor{gray!5}Meta-Llama-3-8B-Instruct & 67.81 & 0.68 & 0.26 & 0.40 & 67.70 & 0.68 & 12.52 & 0.40 & 0.50 \\
\rowcolor{gray!5}glm-4-9b-chat & 71.62 & 0.72 & 1.55 & 0.41 & 77.97 & 0.41 & 33.15 & 0.44 & 0.86  \\
\rowcolor{gray!5}Baichuan2-13B-Chat & 72.50 & 0.73 & 0.00 & 0.42 & 58.83 & 0.58 & 0.00 & 0.37 & 0.00 \\
\rowcolor{gray!5}DeepSeek-V3(671B) & 79.47 & 0.79 & 4.88 & 0.44 & 80.02 & 0.80 & 37.52 & 0.44 & 1.32 \\

\rowcolor{yellow!5}Doubao-Pro & 77.85 & 0.78 & 2.35 & 0.43 & 76.33 & 0.76 & 29.74 & 0.43 & 0.84 \\
\rowcolor{yellow!5}Baichuan3-Turbo & 72.76 & 0.72 & 1.02 & 0.42 & 71.19 & 0.71 & 20.14 & 0.41 & 0.25 \\
\rowcolor{yellow!5}Qwen-Plus & 77.64 & 0.77 & 3.08 & 0.44 & 76.90 & 0.78 & 29.95 & 0.44 & 0.60 \\
\rowcolor{yellow!5}gpt-4-Turbo & 74.02 & 0.74 & 1.17 & 0.42 & 77.26 & 0.77 & 30.53 & 0.43 & 0.42 \\
\rowcolor{yellow!5}gpt-4o & 76.64 & 0.76 & 2.55 & 0.43 & 78.41 & 0.78 & 33.24 & 0.44 & 0.75 \\

\rowcolor{blue!5} \multicolumn{10}{l}{\textbf{\textit{TGMA(ours)}}} \\
\rowcolor{gray!5} DeepSeek-V3(671B) & 80.02 & 0.80 & 6.71 & 0.43 & 80.97 & \textbf{0.87} & 43.90 & \textbf{0.47} & 2.03 \\
\rowcolor{gray!5} Qwen2.5-7B-Instruct(SFT) & \textbf{83.05} & 0.83 & \textbf{8.89} & \textbf{0.62} & 88.78 & 0.86 & \textbf{44.29} & 0.46 & \textbf{4.06} \\
\rowcolor{yellow!5} gpt-4o & 76.90 & \textbf{0.87} & 4.49 & 0.47 & \textbf{89.90} & 0.83 & 43.88 & 0.44 & 2.31 \\

\bottomrule
\end{tabular}
\caption{Travel Plan Evaluation under One-shot In-Context Learning Setting. Pass Rate (PR.) metrics are reported in percentages (\%). Gray and yellow
backgrounds denote open-source and closed-source models.}
\label{tab:additional_plan_comparison_one-shot}
\end{threeparttable}
\end{table*}

\begin{table*}[tp]
\centering
\setlength{\tabcolsep}{11pt}
\renewcommand{\arraystretch}{1.1}
\begin{tabular}{cc|cccc|c|c|c}
\toprule
\multicolumn{2}{c|}{Setting} & \multicolumn{4}{c|}{BLEU} & \multirow{2}{*}{METEOR} & \multirow{2}{*}{ROUGE-L} & \multirow{2}{*}{CIDEr} \\
\cmidrule{1-6}
KB & Topic & B-1 & B-2 & B-3 & B-4 & & & \\
\midrule
 \cmark & \xmark & 9.29 & 2.58 & 0.51 & 0.18 & 19.89 & 16.67 & 0.336 \\
 \xmark & \cmark & 5.42 & 1.49 & 0.32 & 0.12 & 15.67 & 15.82 & 0.335 \\
 \xmark & \xmark & 4.82 & 1.27 & 0.35 & 0.16 & 15.63 & 15.90 & 0.336 \\
 \cmark & \cmark & 10.96 & 4.50 & 2.29 & 1.36 & 22.05 & 16.85 & 0.343 \\
\bottomrule
\end{tabular}
\caption{Ablation study on Decision-Making Evaluation based on our framework with GPT-4o. ROUGE, BLEU, and METEOR metrics are expressed as percentages (\%). \cmark/\xmark indicates whether the corresponding module is used or not.}
\label{table:ablation_study}
\end{table*}

\begin{table*}[tp]
\centering
\setlength{\tabcolsep}{9pt}
\renewcommand{\arraystretch}{1.1}
\begin{tabular}{cc|cccc|cccc|c}
\toprule
\multicolumn{2}{c|}{Setting} & \multicolumn{4}{c|}{Commonsense Constraint} & \multicolumn{4}{c|}{User Preference Constraint} & \multirow{3}{*}{\makecell{Final Pass\\ Rate}} \\
\cmidrule{1-10}
\multirow{2}{*}{ID} & \multirow{2}{*}{OP} & \multicolumn{2}{c}{Micro} & \multicolumn{2}{c|}{Macro} & \multicolumn{2}{c}{Micro} & \multicolumn{2}{c|}{Macro} & \\
\cmidrule{3-10}
& & PR. & F1 & PR. & F1 & PR. & F1 & PR. & F1 & \\
\midrule
 \cmark & \xmark & 47.36 & 0.47 & 3.30 & 0.28 & 49.53 & 0.50 & 27.74 & 0.29 & 0.47  \\
 \xmark & \cmark & 76.70 & 0.71 & 6.80 & 0.44 & 79.67 & 0.79 & 36.80 & 0.44 & 0.67 \\
 \xmark & \xmark & 72.19 & 0.64 & 7.33 & 0.43 & 77.01 & 0.78 & 33.60 & 0.44 & 0.38 \\
 \cmark & \cmark & 78.65 & 0.79 & 33.49 & 0.44 & 80.30 & 0.81 & 37.30 & 0.45 & 1.07 \\
\bottomrule
\end{tabular}
\caption{Ablation study on Travel Plan Evaluation based on our Multi-Agent framework with GPT-4o. Pass Rate (PR.) and F1 metrics are reported in percentages (\%). ID and OP denote Intent Detection and Overall Planning agents respectively.}
\label{tab:plan_comparison}
\end{table*}

Tables \ref{tab:additional_plan_comparison_cot} and \ref{tab:additional_plan_comparison_one-shot} demonstrate the significant improvements achieved by TGMA in travel plan generation. Our enhanced Qwen2.5-7B-Instruct with SFT shows remarkable performance, achieving the highest commonsense constraint micro-precision of 83.39\% in CoT and 83.05\% in one-shot settings, surpassing larger models like gpt-4o. In handling user preferences, our approach pushes the micro-precision to 87.94\% and 88.78\% respectively, marking substantial gains over baseline performances.

Most significantly, our method addresses the critical challenge of macro-precision scores. While baseline models show macro-precision scores near zero for commonsense constraints, our enhanced models achieve scores of 8.12\% and 8.89\% across different settings, demonstrating more consistent performance across varied scenarios. The improvement extends to user preference constraints, with macro-precision scores above 43\% consistently, indicating better handling of diverse user requirements.

The final pass rate metric clearly demonstrates our method's effectiveness. Our enhanced Qwen2.5-7B-Instruct achieves pass rates of 2.99\% and 4.06\%, significantly outperforming both open-source and closed-source baselines. This improvement represents success in simultaneously satisfying both commonsense and user preference constraints, a challenging requirement that most baseline models struggle with, as evidenced by their near-zero pass rates. These results validate TGMA as an effective solution for generating practical and user-aligned travel plans.

\subsection{Ablation Study}

\subsubsection{Decision-Making Support}

As shown in Table \ref{table:ablation_study}, our ablation study reveals the crucial role of both knowledge base and topic-guided components in travel requirement clarification. The full model incorporating both components achieves optimal performance across all metrics, with BLEU-1 reaching 10.96 and BLEU-4 at 1.36, significantly outperforming all ablated variants. The knowledge base component proves particularly vital, as its removal causes BLEU-1 to drop sharply to 5.42, while removing the topic-guided component results in BLEU-1 declining to 9.29. 

The synergistic effect of both components is further evidenced by the METEOR scores, where the full model achieves 22.05, compared to 19.89 with only knowledge base and 15.67 with only topic guidance. The model's performance deteriorates most severely when both components are removed, with BLEU-1 falling to 4.82, demonstrating that neither component alone is sufficient for optimal performance. These results strongly validate our design choice of integrating both knowledge-based reasoning and topic-guided interaction for effective travel requirement clarification.

\begin{table*}[h]
\centering
\setlength{\tabcolsep}{9pt}
\renewcommand{\arraystretch}{1.2}
\begin{tabular}{l|cccc|c|c|c}
\toprule
\multirow{2}{*}{Model} & \multicolumn{4}{c|}{BLEU} & \multirow{2}{*}{METEOR} & \multirow{2}{*}{ROUGE-L} & \multirow{2}{*}{CIDEr} \\
\cmidrule{2-5}
& B-1 & B-2 & B-3 & B-4 & & & \\

\midrule
\rowcolor{blue!5} \multicolumn{8}{c}{\textbf{Pre-fine-tuning Performance}} \\
\midrule
glm-4-9b-chat & 7.54 & 3.90 & 1.94 & 1.18 & 17.09 & 15.14 & 0.348 \\
Qwen2.5-7B-Instruct & 8.29 & 3.23 & 1.44 & 0.79 & 17.26 & 15.34 & 0.343 \\
Baichuan2-13B-Chat & 9.49 & 4.41 & 1.76 & 0.96 & 14.74 & 15.49 & 0.345 \\

\midrule
\rowcolor{blue!5} \multicolumn{8}{c}{\textbf{Post-fine-tuning Performance}} \\
\midrule
glm-4-9b-chat & 14.41 & 7.22 & 3.79 & 2.35 & 18.84 & 16.57 & 0.353 \\
Qwen2.5-7B-Instruct & 12.89 & 6.02 & 3.02 & 1.79 & 17.67 & 16.19 & 0.360 \\
Baichuan2-13B-Chat & 10.28 & 3.86 & 1.72 & 0.92 & 18.64 & 15.89 & 0.347 \\

\bottomrule
\end{tabular}
\caption{Decision-Making Support Evaluation comparing model performance before and after fine-tuning. ROUGE, BLEU, and METEOR metrics are expressed as percentages (\%).}
\label{table:finetune_comparison}
\end{table*}

\begin{table*}[!htb]
\normalsize
\setlength{\tabcolsep}{7pt}
\renewcommand{\arraystretch}{1.2}
\centering
\begin{tabular}{l|cccc|cccc|c}
\toprule
\multirow{4}{*}{Model} & \multicolumn{4}{c|}{Commonsense Constraint} & \multicolumn{4}{c|}{User Preference Constraint} & \multirow{4}{*}{\makecell{Final Pass\\Rate}} \\
\cmidrule(lr){2-5} \cmidrule(lr){6-9}
& \multicolumn{2}{c}{Micro} & \multicolumn{2}{c|}{Macro} & \multicolumn{2}{c}{Micro} & \multicolumn{2}{c|}{Macro} & \\
\cmidrule(lr){2-3} \cmidrule(lr){4-5} \cmidrule(lr){6-7} \cmidrule(lr){8-9}
& PR. & F1 & PR. & F1 & PR. & F1 & PR. & F1 & \\
\midrule
\rowcolor{blue!5} \multicolumn{10}{c}{\textbf{Pre-fine-tuning Performance}} \\
\midrule

glm-4-9b-chat & 54.13 & 0.54 & 0.17 & 0.35 & 77.32 & 0.77 & 29.69 & 0.43 & 0.10 \\
Qwen2.5-7B-Instruct & 49.38 & 0.50 & 0.00 & 0.32 & 74.49 & 0.74 & 22.87 & 0.42 & 0.00 \\
Baichuan2-13B-Chat & 50.00 & 0.50 & 0.00 & 0.33 & 60.42 & 0.75 & 0.00 & 0.43 & 0.00 \\

\midrule
\rowcolor{blue!5} \multicolumn{10}{c}{\textbf{Post-fine-tuning Performance}} \\
\midrule

glm-4-9b-chat & 71.50 & 0.72 & 2.20 & 0.41 & 75.17 & 0.42 & 38.33 & 0.44 & 1.00 \\
Qwen2.5-7B-Instruct & 73.71 & 0.73 & 1.57 & 0.42 & 85.55 & 0.86 & 56.60 & 0.46 & 1.40 \\
Baichuan2-13B-Chat & 65.82 & 0.66 & 0.95 & 0.38 & 71.43 & 0.72 & 32.15 & 0.42 & 0.72 \\

\bottomrule
\end{tabular}
\caption{Travel Plan Evaluation comparing model performance before and after fine-tuning. Pass Rate (PR.) metrics are reported in percentages (\%).}
\label{tab:plan_comparison}
\end{table*}

\subsubsection{Travel Plan Generation}

As shown in Table \ref{tab:plan_comparison}, our ablation study demonstrates the effectiveness of both Intent Detection (ID) and Overall Planning (OP) agents in travel plan generation. The full model with both modules achieves the best overall performance, with micro-precision reaching 78.65\% for commonsense constraints and 80.30\% for user preferences, along with a final pass rate of 1.07\%. The OP module proves particularly crucial, as its removal causes significant performance degradation across all metrics, most notably in commonsense constraint micro-precision dropping to 47.36\% and user preference micro-precision falling to 49.53\%.

Most significantly, the macro-precision scores reveal the complementary nature of these modules. The full model achieves 33.49\% macro-precision for commonsense constraints, substantially outperforming variants with single or no modules, which only achieve scores below 8\%. This dramatic improvement in macro-precision indicates our complete framework's superior ability to handle diverse travel planning scenarios consistently. These results validate our design choice of combining constraint-aware intent detection with controlled planning for robust travel plan generation.

\subsection{Fine-tuning Analysis}

\subsubsection{Decision-Making Support}

As shown in Table \ref{table:finetune_comparison}, our Dialog-SFT approach demonstrates significant improvements in models' ability to clarify travel requirements. The most notable enhancement is observed in glm-4-9b-chat, where BLEU-4 scores increase from 1.18 to 2.35, accompanied by substantial gains across all metrics, with BLEU-1 improving from 7.54 to 14.41 and METEOR rising from 17.09 to 18.84. Similarly, Qwen2.5-7B-Instruct shows remarkable progress with BLEU-4 more than doubling from 0.79 to 1.79.

Interestingly, smaller models achieve more substantial improvements compared to larger ones, as evidenced by Baichuan2-13B-Chat's relatively modest gains. This suggests our Dialog-SFT approach effectively enhances model capabilities regardless of model size, particularly benefiting smaller architectures. The consistent improvements in METEOR and ROUGE-L scores across all models indicate enhanced semantic understanding and response coherence, validating the effectiveness of our fine-tuning strategy in improving travel requirement clarification capabilities.

\subsubsection{Travel Plan Generation}

As shown in Table \ref{tab:plan_comparison}, our fine-tuning approach demonstrates substantial improvements in models' travel planning capabilities. Qwen2.5-7B-Instruct shows the most remarkable enhancement, with its micro-precision for commonsense constraints increasing from 49.38\% to 73.71\%, and user preference constraints improving significantly from 74.49\% to 85.55\%. Most notably, its macro-precision for user preferences sees a dramatic increase from 22.87\% to 56.60\%, indicating greatly improved consistency across diverse planning scenarios.

The effectiveness of our fine-tuning approach is further validated by the final pass rate improvements. All models progress from near-zero pass rates to achieving meaningful success, with Qwen2.5-7B-Instruct reaching 1.40\%. The consistent improvements in both commonsense and user preference constraints across all models, particularly in macro-precision metrics, demonstrate our fine-tuning strategy's effectiveness in enhancing models' ability to generate travel plans that satisfy multiple practical constraints simultaneously.

\section{Tourism Knowledge Base Examples}
\label{appendix B}
We present examples from our tourism knowledge base, which covers five main categories: attractions, accommodations, dining, transportation, and weather information. These examples demonstrate the structured information used to support our dialogue system.

\subsection{City}
Our tourism dataset encompasses 24 major cities across China, representing a diverse geographic and cultural landscape. These cities span from the northern regions (Beijing, Tianjin, Harbin, Shenyang, and Dalian), through the eastern coastal areas (Shanghai, Nanjing, Hangzhou, Suzhou, and Xiamen), to the southern region (Guangzhou, Shenzhen, and Fuzhou). The dataset also covers central China (Wuhan, Changsha, and Nanchang), western China (Chengdu, Xi'an, Lanzhou, and Xining), and southwestern China (Kunming, Guiyang, and Chongqing). This comprehensive coverage ensures our dataset captures the rich diversity of China's tourism resources, from ancient historical sites to modern urban attractions, and from coastal sceneries to inland landscapes. These cities were selected based on their tourism significance, economic importance, and distinctive cultural characteristics, providing a representative sample of China's major tourist destinations.

\begin{figure}[htbp]
    \centering
    \includegraphics[width=0.5\textwidth]{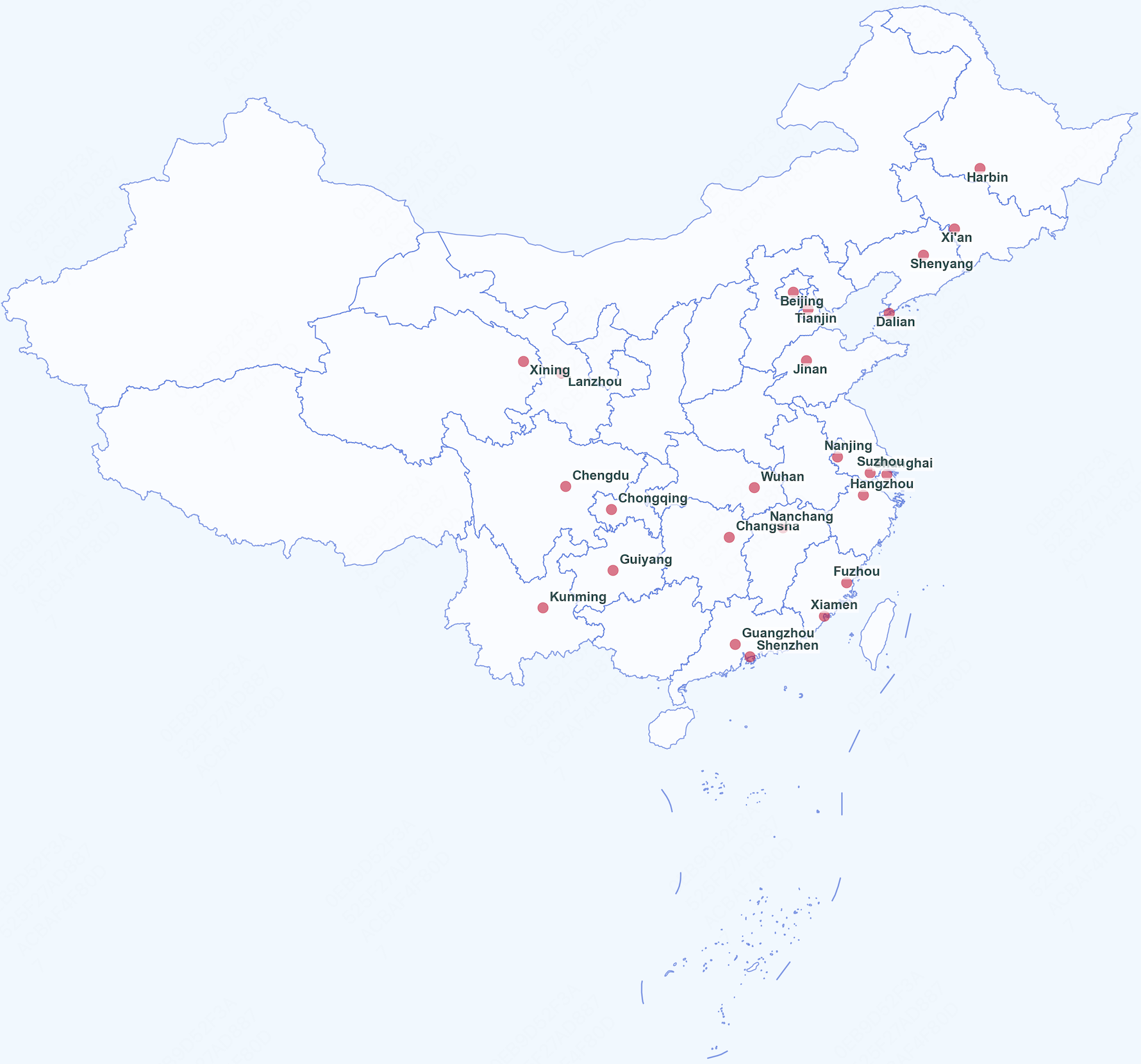}
    \caption{Geographic distribution of the 24 cities in our dataset}
\end{figure}

\subsection{Attraction}

\begin{table*}[tp]
\centering
\small
\begin{tcolorbox}[
    width=1.0\textwidth,
    colback=lightgray,    
    colframe=black,
    arc=3mm,
    boxrule=0.5pt
]
\begin{CJK}{UTF8}{gbsn}\textbf{Attraction name:} Jiufeng Forest Zoo

\textbf{Introduction:} Wuhan Jiufeng Forest Zoo, nestled in Jiufeng National Forest Park, is just 12km from downtown Wuhan. Set across 53.3 hectares with 85\% forest coverage, the zoo houses diverse wildlife in specialized areas including tiger, lion, bear, and monkey gardens. The facility features an animal kindergarten, a parent-child eco-park, and two popular performance venues - the "Dream Theater" and "Bear and Tiger Gathering."

\textbf{Basic Information:}
\begin{itemize}[noitemsep,topsep=0pt]
\item Image: \includegraphics[scale=0.035]{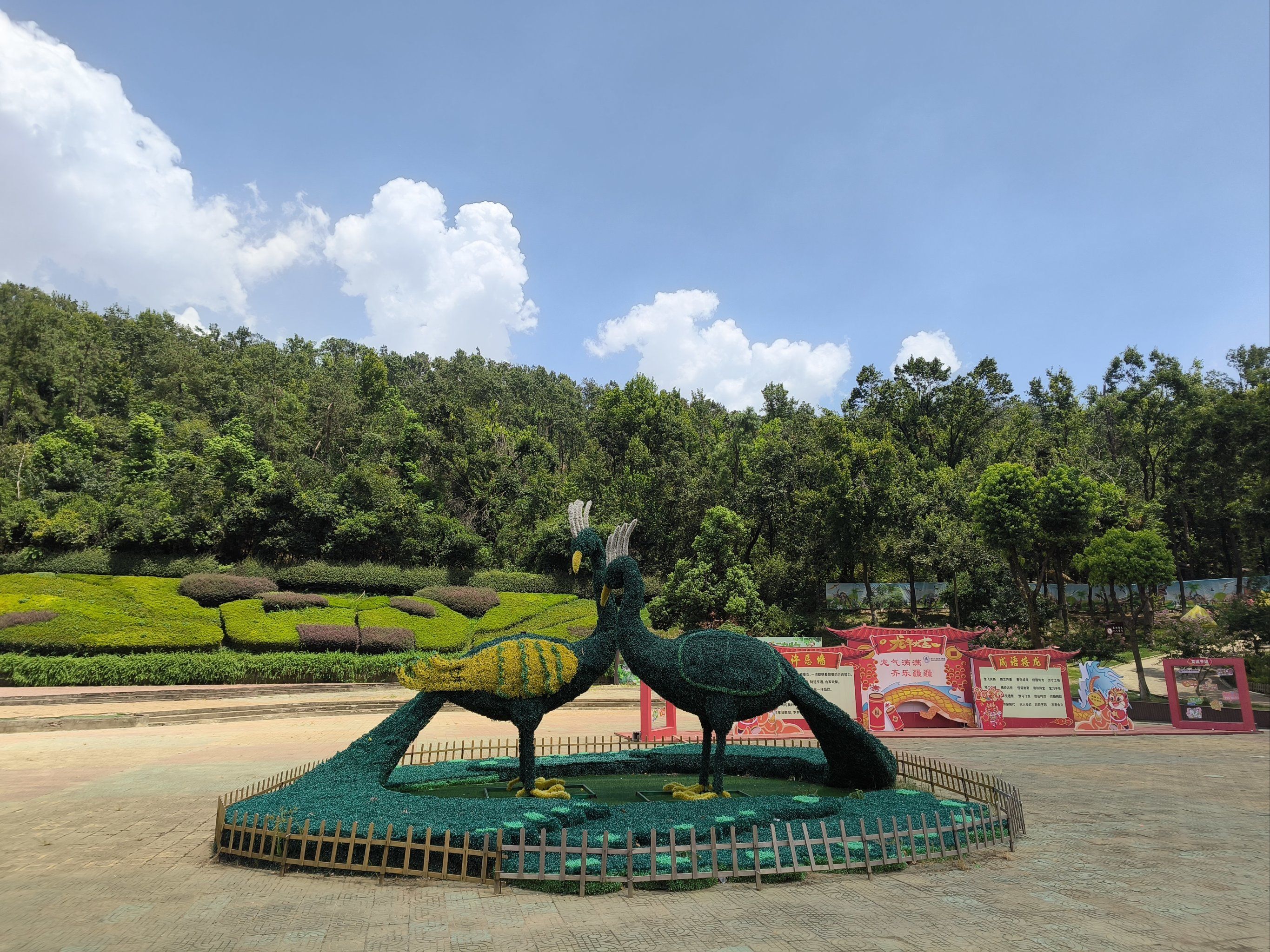}

\item Address: No. 1 Special, Jiufeng Lion Peak, Hongshan District, Wuhan City, Hubei Province
\item Phone: 027-87639763/027-87639013
\item Geographic Coordinates: 114.49225° East, 30.500436° North
\item Type: Attraction-Zoo-Botanical and Zoological Garden
\end{itemize}

\textbf{Visitor Information:}
\begin{itemize}[noitemsep,topsep=0pt]
\item Best Season to Visit: Suitable for all seasons
\item Recommended Visit Duration: 3-5 hours
\item Must-Visit Index: Must-see attraction
\item Average Expense per Person: 83 yuan
\item Rating: 3.8 stars
\end{itemize}

\textbf{Nearby Restaurants:}
\begin{itemize}[noitemsep,topsep=0pt]
\item Green Tea Restaurant (Wuhan Joy City Branch) - 1.5km away
\item Xiaofu Nan Hunan Cuisine (Joy City Branch) - 1.5km away
\item Victoria's Style Original Cut Steak Western Restaurant (Joy City Branch) - 1.5km away
\end{itemize}

\textbf{Nearby Accommodations:}
\begin{itemize}[noitemsep,topsep=0pt]
\item \textit{Chain Hotels:}
  \begin{itemize}[noitemsep,topsep=0pt]
  \item City Comfort Inn (Wuhan Jiufeng Mountain Forest Park Branch) - 1.0km away
  \item Fengyi Fashion Hotel (Wuhan Jiufeng Mountain Forest Park Branch) - 0.9km away
  \item City Comfort Inn (Wuhan Optics Valley Science and Technology Convention Center Branch) - 2.1km away
  \end{itemize}
\item \textit{Upscale Hotels:}
  \begin{itemize}[noitemsep,topsep=0pt]
  \item Yishang Hotel (Wuhan Optics Valley Bio-city Joy City Branch) - 0.9km away
  \item Wyndham Garden Wuhan Optics Valley - 1.9km away
  \item Yishang Hotel (Wuhan Optics Valley Science and Technology Convention Center Joy City Branch) - 2.2km away
  \end{itemize}
\end{itemize}

\textbf{Visitor Reviews:}
\begin{itemize}[noitemsep,topsep=0pt]
\item The zoo boasts extensive forest coverage and diverse wildlife, with family-friendly attractions including a children's farm and interactive areas.

\item Located just 40 minutes from downtown, the zoo offers convenient access and parking. A typical visit takes 2-3 hours, featuring midday animal performances.

\item Interactive animal feeding experiences are available for visitors seeking closer encounters with wildlife.
\end{itemize}

\textbf{Ticket Information:}
\begin{itemize}[noitemsep,topsep=0pt]
\item \textit{Single Admission Ticket:}
  \begin{itemize}[noitemsep,topsep=0pt]
  \item Adult Ticket: 55-60 yuan
  \item College Student Ticket: 36-40 yuan
  \end{itemize}
\item \textit{Package Deals:}
  \begin{itemize}[noitemsep,topsep=0pt]
  \item Two-Person Ticket: 108-119 yuan
  \item Family Ticket (1 Adult + 1 Child): 85 yuan
  \item Family Ticket (2 Adults + 1 Child): 138 yuan
  \end{itemize}
\item \textit{Special Attraction Tickets:}
  \begin{itemize}[noitemsep,topsep=0pt]
  \item Forest Secret Realm (Immersive Holographic Interaction): 19-20 yuan
  \item Family Farm + Forest Secret Realm Package: 29.9 yuan
  \end{itemize}
\item \textit{Combination Offers:}
  \begin{itemize}[noitemsep,topsep=0pt]
  \item Adult Ticket + Forest Secret Realm: 77 yuan
  \item College Student Ticket + Forest Secret Realm: 57 yuan
  \end{itemize}
\end{itemize}
\end{CJK}
\end{tcolorbox}
\caption{A detailed attraction entry showcasing comprehensive attraction information spanning location details, visitor guidance, nearby POIs (restaurants and accommodations), ticket options, and user feedback for Jiufeng Forest Zoo.}
\label{tab:attraction example}
\end{table*}

Table~\ref{tab:attraction example} presents a detailed example of an attraction entry from our tourism database. The entry encompasses comprehensive basic information including name, introduction, visual content, location, and contact details. It also contains visitor-oriented practical information such as recommended visiting times and costs, along with surrounding service details covering nearby restaurants and accommodations within 3km. Additionally, the entry includes visitor reviews providing qualitative insights and detailed ticket information with various pricing options. This structured format is consistently applied across our database, ensuring a holistic representation of each attraction through both factual information and user-generated content.

\subsection{Restaurant}

\begin{table*}[tp]
\centering
\small
\begin{tcolorbox}[
    width=1.0\textwidth,
    colback=lightgray,    
    colframe=black,
    arc=3mm,
    boxrule=0.5pt
]
\begin{CJK}{UTF8}{gbsn}
\textbf{Restaurant name:} Xiabu Xiabu (Wangfujing Department Store)\\
\textbf{Introduction:} Established in Beijing in 1998, Xiabu Xiabu pioneered counter-style hotpot dining with various soup bases (clear, spicy, sour-spicy, curry) and signature sesame sauce. The restaurant's popularity is reflected in its constant queues, though efficient service ensures short waiting times.\\
\textbf{Basic Information:}
\begin{itemize}[noitemsep,topsep=0pt]
\item Image: \includegraphics[scale=0.05]{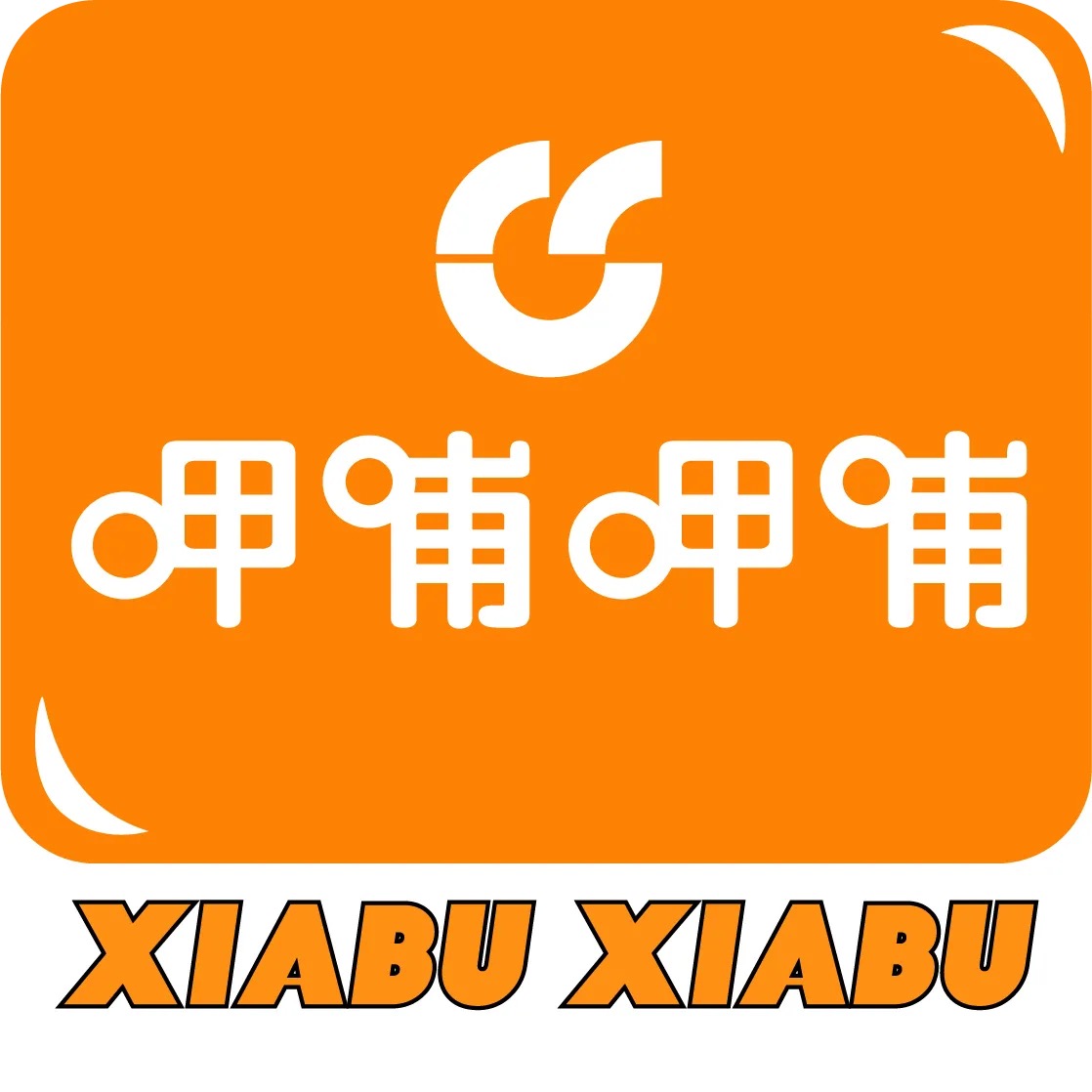}
\item Address: 7th Floor, North Building, Wangfujing Department Store, 255 Wangfujing Street, Dongcheng District, Beijing
\item Phone: 010-85295567
\item Geographic Coordinates: 116.410505° East, 39.914053° North
\item Type: Individual Hotpot
\item Opening Hours: 10:00-21:00 (Monday-Sunday)
\item Parking: Free parking available
\item Average Price: 64 yuan
\item Rating: 4.5 stars
\end{itemize}

\textbf{Reviews:}
\begin{itemize}[noitemsep,topsep=0pt]
\item I frequently bring my children here. The environment is decent, and the customer flow is moderate, making it comfortable for family dining.

\item I always choose Xiabu Xiabu when dining out - it's economical and satisfying.

\item The ingredients are fresh, and the butter-based spicy soup base is wonderfully aromatic. The Cloud Peach Jasmine tea has a perfect sweetness level. Although we ate in a hurry, the overall experience was excellent.
\end{itemize}

\end{CJK}
\end{tcolorbox}
\caption{A detailed restaurant entry demonstrating comprehensive dining information including basic details, operational hours, dining environment, and customer reviews for Xiabu Xiabu.}
\label{tab:restaurant example}
\end{table*}

Table~\ref{tab:restaurant example} presents a detailed restaurant entry from our tourism database. The entry provides comprehensive information about Xiabu Xiabu, a popular hotpot restaurant in Beijing's Wangfujing area. It includes basic operational details, location information, and customer reviews. The data structure encompasses both objective information (such as opening hours and pricing) and subjective feedback through customer reviews, offering potential visitors a complete picture of the dining experience.

\subsection{Hotel}

\begin{table*}[tp]
\centering
\small
\begin{tcolorbox}[
    width=1.0\textwidth,
    colback=lightgray,    
    colframe=black,
    arc=3mm,
    boxrule=0.5pt
]
\begin{CJK}{UTF8}{gbsn}
\textbf{Hotel name:} Shanghai Zunmao Hotel

\textbf{Introduction:} Located in Lujiazui Financial District, this tech-savvy hotel offers rooms with internet connectivity and on-demand entertainment. Featuring both Chinese and Western restaurants plus a modern bar, the hotel provides comprehensive recreational facilities, conference spaces and banquet halls. Recently renovated floors ensure enhanced guest comfort. Our staff looks forward to serving you!

\textbf{Basic Information:}
\begin{itemize}[noitemsep,topsep=0pt]
\item Image: \includegraphics[scale=0.05]{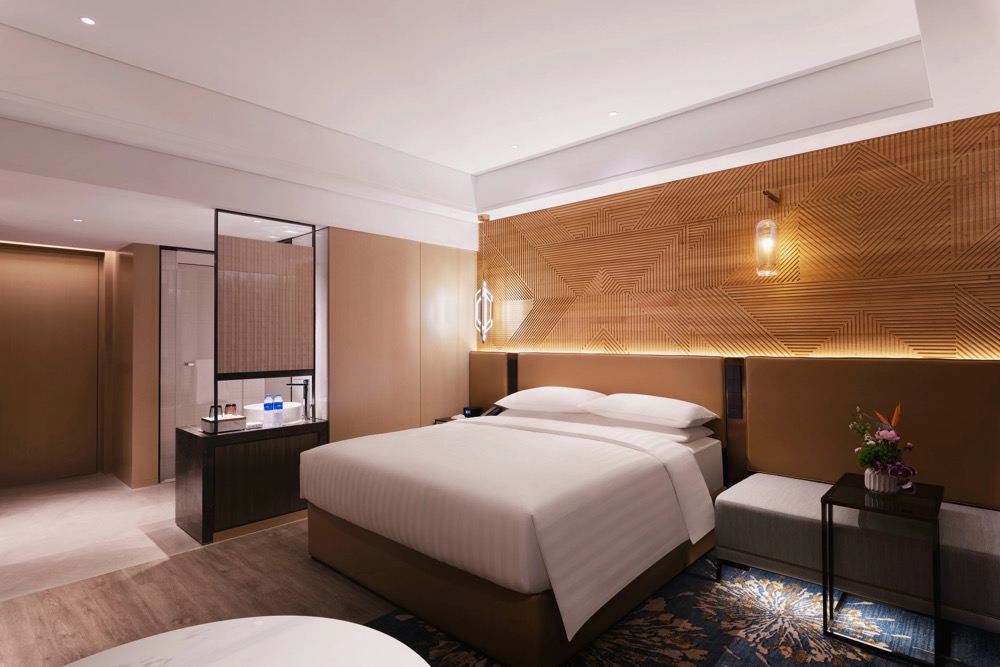}
\item Address: 357 Songlin Road, Pudong New Area, Shanghai
\item Phone: 021-58300000
\item Geographic Coordinates: 121.535635° East, 31.226312° North
\item Category: Four-Star/Luxury
\end{itemize}

\textbf{Hotel Information:}
\begin{itemize}[noitemsep,topsep=0pt]
\item Parking: First 3 hours free
\item WiFi: Available
\item Stars: 4.5 stars
\end{itemize}

\textbf{Room Types and Rates:}
\begin{itemize}[noitemsep,topsep=0pt]
\item Superior King Room: 608 yuan (Entry-level option)
\item Deluxe Twin Room: 717 yuan (Mid-range option)
\item Executive King Room: 1168 yuan (Business-class option)
\item Business Suite: 1672 yuan (Premium option)
\item Duplex Executive Apartment: 2888 yuan (Luxury option)
\end{itemize}

\textbf{Guest Reviews:}
\begin{itemize}[noitemsep,topsep=0pt]
\item Manager Xiao Luo delivers exceptional service with great patience and efficiency.


\item A well-established hotel near the Stock Exchange in Pudong. Recently renovated and maintains high standards for business conferences.

\item Clean, spacious rooms with modern renovations. Front desk staff member Xu Jiahui provides warm and attentive service.

\end{itemize}
\end{CJK}
\end{tcolorbox}
\caption{A detailed hotel entry illustrating detailed accommodation information spanning location details, facilities, room categories with pricing, and guest reviews for Shanghai Zunmao Hotel.}
\label{tab:hotel example}
\end{table*}

Table~\ref{tab:hotel example} presents a comprehensive hotel entry from our tourism database, showcasing the Shanghai Zunmao Hotel in the Lujiazui Financial District. The entry encompasses detailed information including basic hotel details, facilities, room types with pricing, and guest reviews. This structured format provides potential visitors with both essential operational information and real guest experiences, facilitating informed accommodation decisions.

\subsection{Transportation}

\begin{table*}[tp]
\small
\centering
\begin{minipage}{0.48\textwidth}
\centering
\begin{tcolorbox}[
    colback=lightgray,
    colframe=black,
    arc=3mm,
    boxrule=0.5pt
]
\begin{CJK}{UTF8}{gbsn}\textbf{Origin City:} Hangzhou \\
\textbf{Destination City:} Shanghai\\
\textbf{Departure Station:} Hangzhou South\\
\textbf{Arrival Station:} Shanghai Hongqiao\\
\textbf{Train Number:} G7382\\
\textbf{Train Type:} G-High Speed Rail\\
\textbf{Journey Duration:} 1h 13m\\
\textbf{Departure Time:} 22:30\\
\textbf{Arrival Time:} 23:43\\
\textbf{Ticket Price:} 60.0 yuan
\end{CJK}
\end{tcolorbox}
\caption{A transportation database entry showing detailed high-speed rail journey information including route details, timing, duration, and pricing for travel between Hangzhou and Shanghai.}
\label{tab:transportation example}
\end{minipage}
\hfill
\begin{minipage}{0.48\textwidth}
\centering
\begin{tcolorbox}[
    colback=lightgray,
    colframe=black,
    arc=3mm,
    boxrule=0.5pt
]
\begin{CJK}{UTF8}{gbsn}
\textbf{City:} Beijing\\
\textbf{Date:} 2024-01-01\\
\textbf{High Temperature:} 2°C\\
\textbf{Low Temperature:} -7°C\\
\textbf{Weather:} Cloudy\\
\textbf{Wind:} West wind level 1\\
\textbf{Air Quality:} 103 (Mild pollution)
\end{CJK}
\end{tcolorbox}
\caption{A weather database entry displaying comprehensive environmental conditions including temperature range, weather status, wind conditions, and air quality index for Beijing on January 1, 2024.}
\label{tab:weather example}
\end{minipage}
\end{table*}

Table~\ref{tab:transportation example} shows a sample train service entry from our tourism database. The entry includes key travel information such as station details, schedule, journey duration, and pricing for high-speed rail service between Hangzhou and Shanghai.

\subsection{Weather}

Table~\ref{tab:weather example} demonstrates a weather data entry from our tourism database. The entry includes essential meteorological information such as temperature, weather conditions, wind status, and air quality index, providing tourists with crucial environmental data for their travel planning.

\section{Annotations and Examples of RETAIL}
\label{appendix C}

\subsection{Detailed Dataset Annotations}

\subsubsection{Step 1: Tourism Knowledge Base Validation}
\begin{itemize}[leftmargin=*, noitemsep]
    \item \textbf{Data Cleaning \& Completeness Check}: Thoroughly clean the tourism knowledge base by removing any entries with missing or incomplete key fields, such as POI (Point of Interest) names, locations, operating hours, or accessibility details. Ensure all essential information is standardized and formatted consistently.
    
    \item \textbf{Privacy Protection \& Anonymization}: Handle sensitive data, such as visitor records or booking details, by implementing random replacement algorithms for personal identifiers, contact information, and payment records to comply with data protection regulations.
    
    \item \textbf{Transportation Data Verification \& Optimization}: Validate transportation data by cross-referencing actual ticket prices, schedules, and route availability with official sources. Where direct routes are unavailable, supplement the database with optimized transfer options. Additionally, verify real-time updates for seasonal route changes or service disruptions.
\end{itemize}
This ensures the knowledge base remains accurate, secure, and user-friendly for travelers.

\subsubsection{Step 2: Decision-Making Support Refinement}
\begin{itemize}[leftmargin=*, noitemsep]
    \item \textbf{Dialogue Evaluation}: Systematically evaluate conversations across key quality dimensions, filtering out dialogues exhibiting misaligned recommendations, ignored user needs, redundant suggestions, unnatural transitions, mechanical responses, inconsistent styles, or insufficient clarifications.
    
    \item \textbf{Dialogue Optimization}: Enhance interaction quality by refining prompts and improving dialogue flow to create more natural, human-like travel planning conversations.
\end{itemize}
This refinement process guarantees the system delivers coherent, personalized travel advice while maintaining conversational naturalness comparable to professional travel agents.

\subsubsection{Step 3: Plan Generation and Revision Verification}
\begin{itemize}[leftmargin=*, noitemsep]
    \item \textbf{Feasibility Check}: Verify all generated travel plans for practical viability, eliminating those with unrealistic schedules, incompatible accommodations, unworkable transportation arrangements, or logically conflicting activities.
    
    \item \textbf{Plan Revision}: During modifications, maintain equilibrium between attraction adjustments and transportation coordination, ensuring all revised plans retain both feasibility and alignment with user preferences.
\end{itemize}
This verification process guarantees that all recommended travel plans are practical and tailored to individual needs.

\subsection{Statistics of RETAIL}

\begin{table}[tp]
\centering
\small
\setlength{\tabcolsep}{5pt}
\renewcommand{\arraystretch}{1.3}
\begin{tabular}{l|cccc}
\toprule
\rowcolor{gray!10} \textbf{Type} & \textbf{Total} & \textbf{Training} & \textbf{Val} & \textbf{Test} \\
\midrule
Single-turn & 3,500 & 2,100 & 830 & 700 \\
\rowcolor{gray!5} Single-turn + Revision & 2,500 & 1,500 & 119 & 500 \\
\midrule
Multi-turn & 2,500 & 1,500 & 1,167 & 500 \\
\rowcolor{gray!5} Multi-turn + Revision & 1,500 & 900 & 66 & 300 \\
\bottomrule
\end{tabular}
\caption{\textbf{Distribution of Different Planning Cases.} RETAIL is divided into four categories based on conversation turns and revision requirements. The numbers represent the sample size for each split (Total/Training/Validation/Test).}
\label{tab:planning_distribution}
\end{table}

To systematically evaluate our proposed approach, we construct a comprehensive dataset consisting of both single-turn and multi-turn planning scenarios. As shown in Table~\ref{tab:planning_distribution}, the dataset contains 10,000 cases in total, covering four different types of planning scenarios. Specifically, we collect 3,500 single-turn cases and 2,500 single-turn cases with revision requirements, which represent basic planning scenarios and their modifications respectively. For more complex scenarios, we include 2,500 multi-turn conversation cases and 1,500 multi-turn cases with revision requirements. The dataset is split into training, validation and test sets with approximate ratios of 60\%, 20\% and 20\% respectively, ensuring sufficient samples for model training while maintaining reliable evaluation. This diverse distribution allows us to comprehensively assess the model's capability in handling various planning scenarios with different complexity levels.

\subsection{Case Study on RETAIL}
Our travel planning dataset encompasses four distinct interaction patterns, demonstrating various levels of complexity in travel assistance scenarios.

\subsubsection{Single-Turn Travel Planning}
The dataset contains examples where comprehensive itineraries are generated from single, detailed user requests. These responses demonstrate the ability to process multiple constraints simultaneously, transforming them into structured daily schedules with precise timing, transportation arrangements, accommodation details, and activity recommendations. The planning maintains both practical feasibility and user preference alignment.

\begin{figure*}[tp]
\centering
\includegraphics[width=\textwidth]{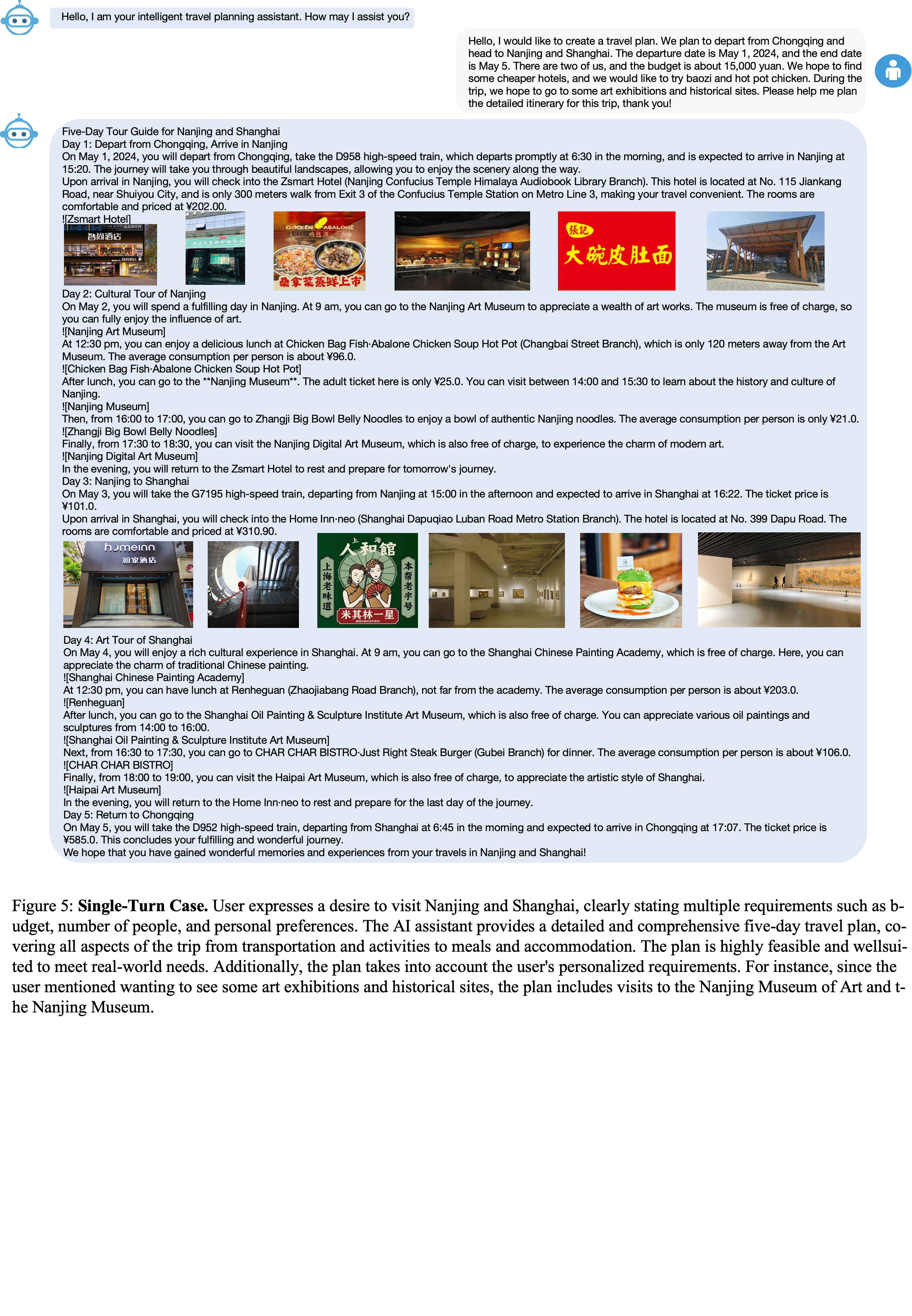}
\end{figure*}

\subsubsection{Single-Turn Planning with Revision}
This category showcases the model's adaptability in real-time plan modification. When presented with new constraints such as weather changes or timing conflicts, the responses demonstrate intelligent adaptation by suggesting appropriate alternative activities while maintaining the original schedule structure and user preferences. The revisions preserve both practical feasibility and the overall travel experience quality.

\begin{figure*}[t]
\centering
\includegraphics[width=\textwidth]{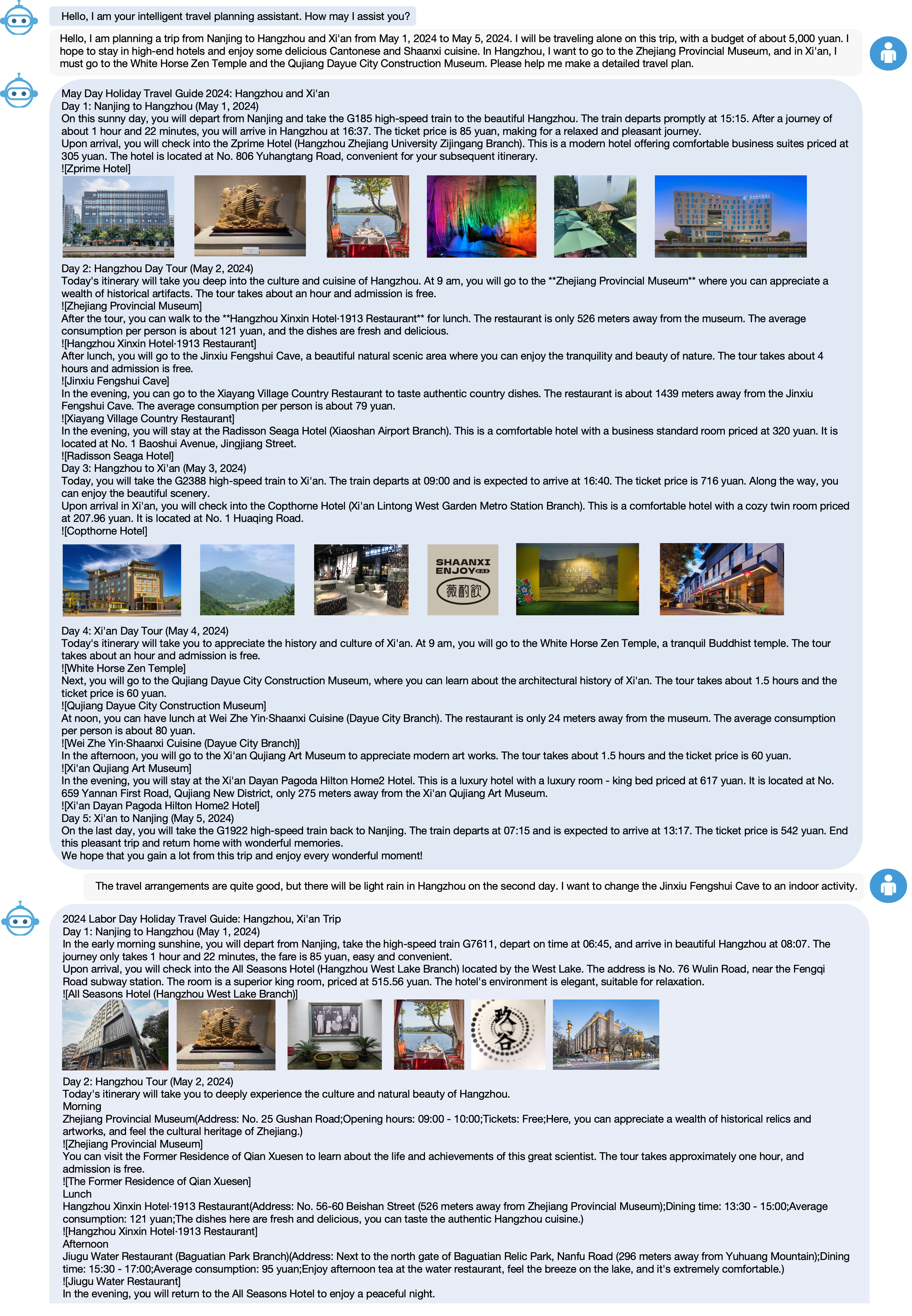}
\end{figure*}
\begin{figure*}[t]
\centering
\includegraphics[width=\textwidth]{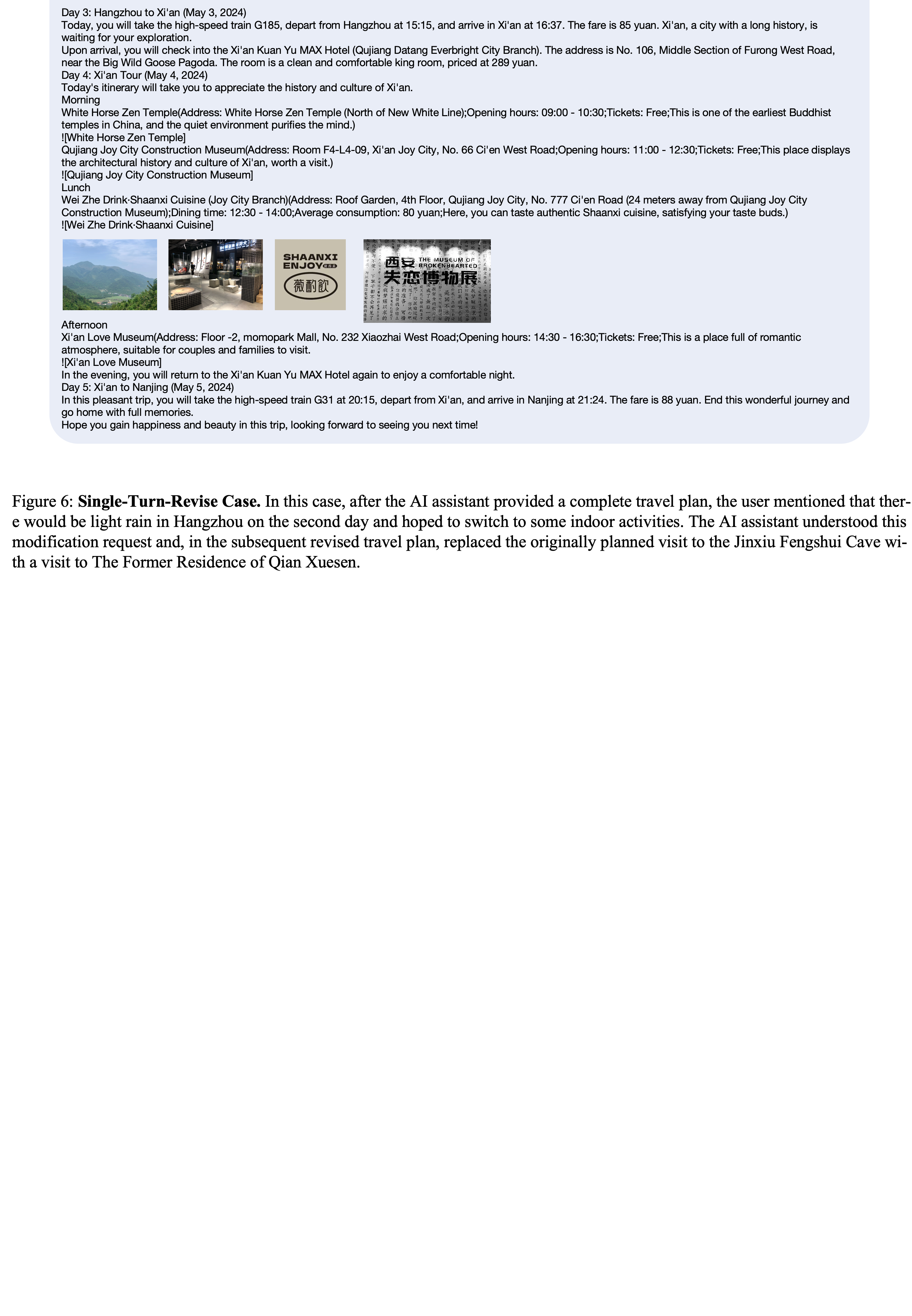}
\end{figure*}

\subsubsection{Multi-Turn Travel Planning}
 These examples illustrate systematic information gathering through progressive dialogue. The model demonstrates sophisticated conversation management by sequentially clarifying essential details about dates, budget, group composition, accommodation preferences, and activity interests. This methodical approach ensures the final itinerary precisely matches user needs while maintaining practical travel logistics.

\begin{figure*}[t]
\centering
\includegraphics[width=\textwidth]{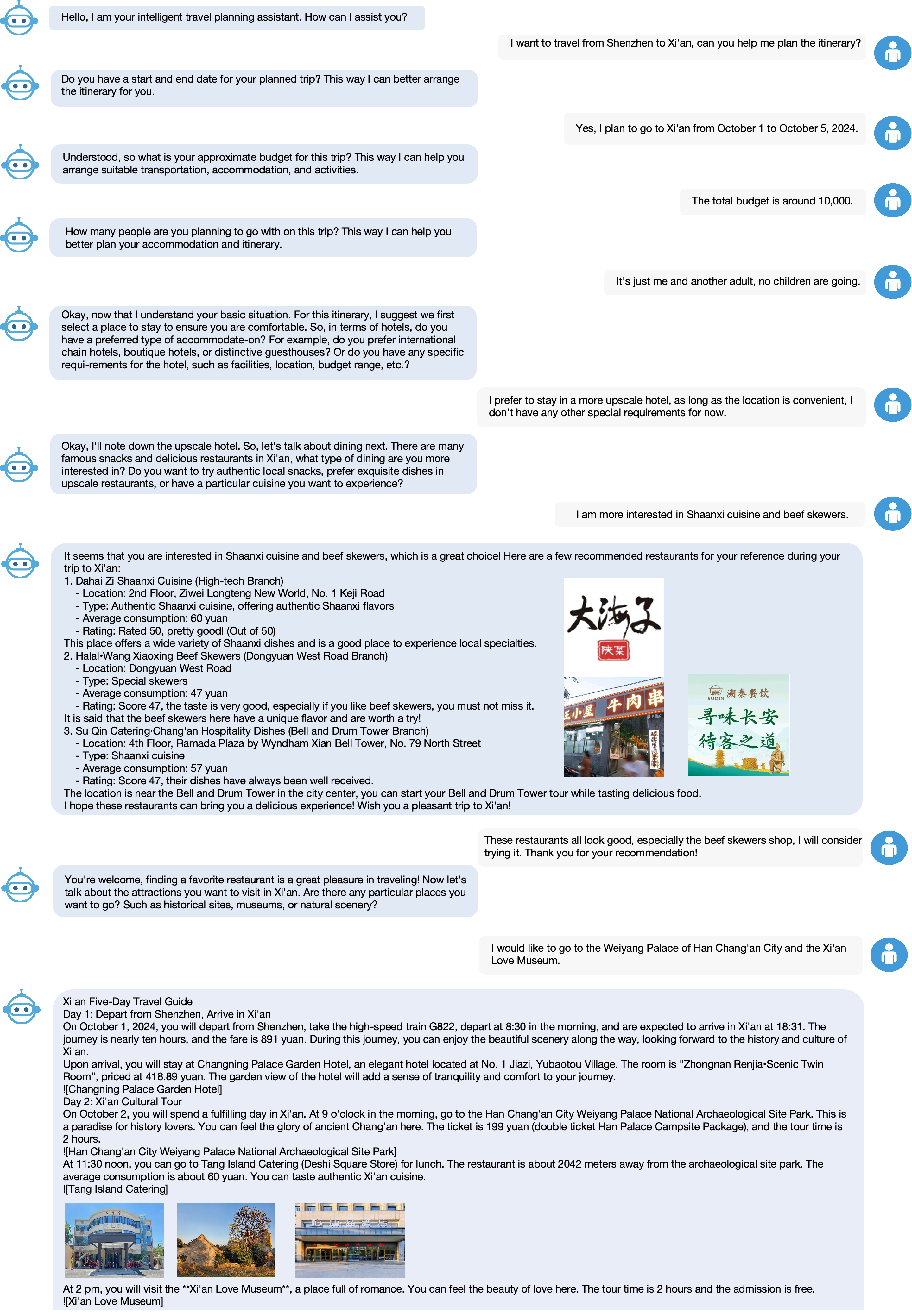}
\end{figure*}
\begin{figure*}[t]
\centering
\includegraphics[width=\textwidth]{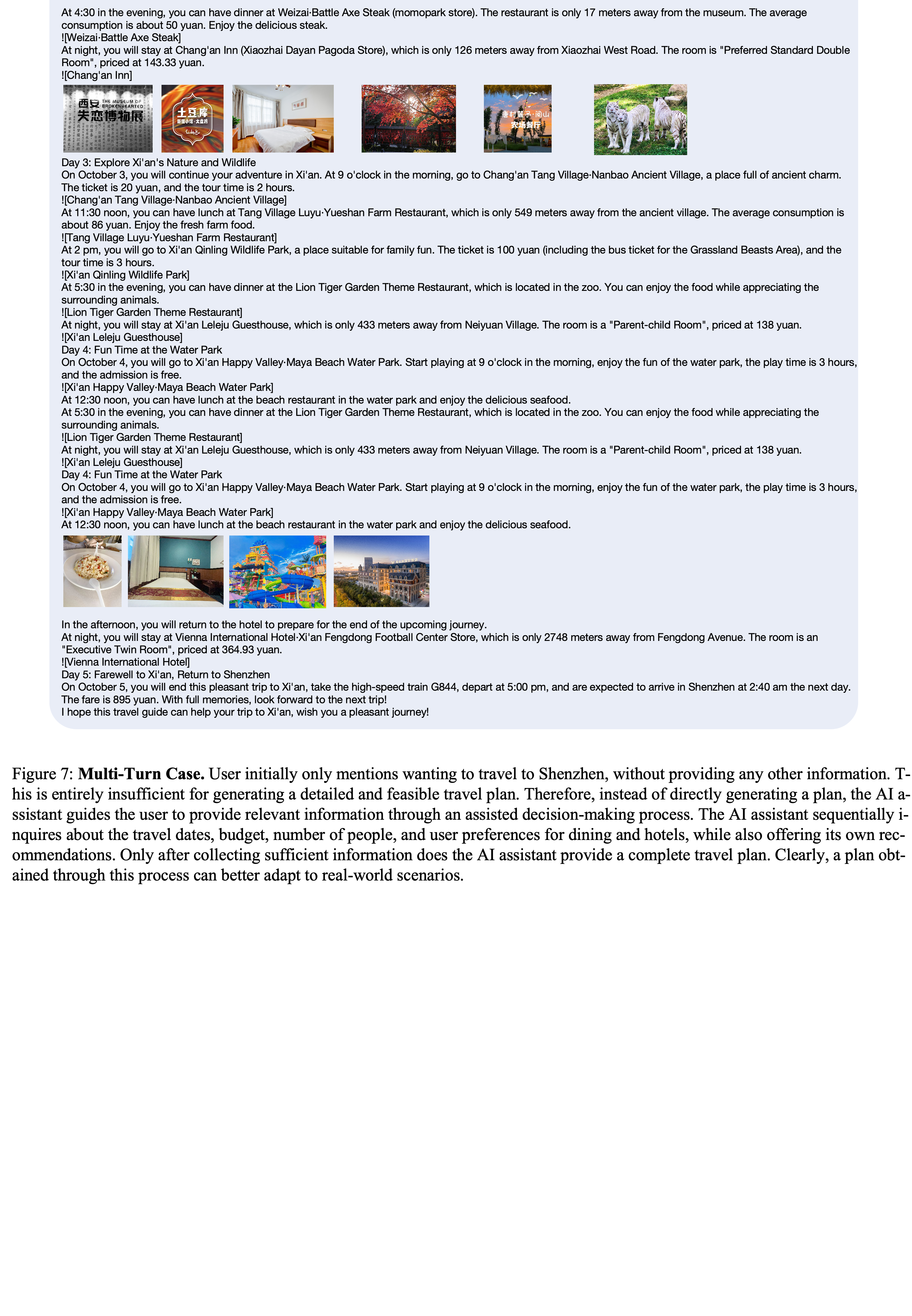}
\end{figure*}

\subsubsection{Multi-Turn Planning with Revision}
The most complex category combines thorough requirement clarification with dynamic plan adaptation. These interactions begin with systematic information gathering through dialogue, followed by itinerary generation and subsequent refinement based on new constraints. This demonstrates the dataset's capability to handle realistic travel planning scenarios requiring both detailed preference understanding and flexible plan modification.

As shown in the following example, the user initially provided only limited information such as the destination and budget. The AI assistant then gradually collected other information through an assisted decision-making process. Once sufficient information was gathered, it provided a complete travel plan. However, the user suggested a modification to the first plan, specifically wanting to replace the Panda Pedestrian Street attraction. After responding to this request, the AI assistant provided a second complete plan.This process clearly demonstrates a procedure that simultaneously incorporates assisted decision-making and plan revision.

\begin{figure*}[t]
\centering
\includegraphics[width=\textwidth]{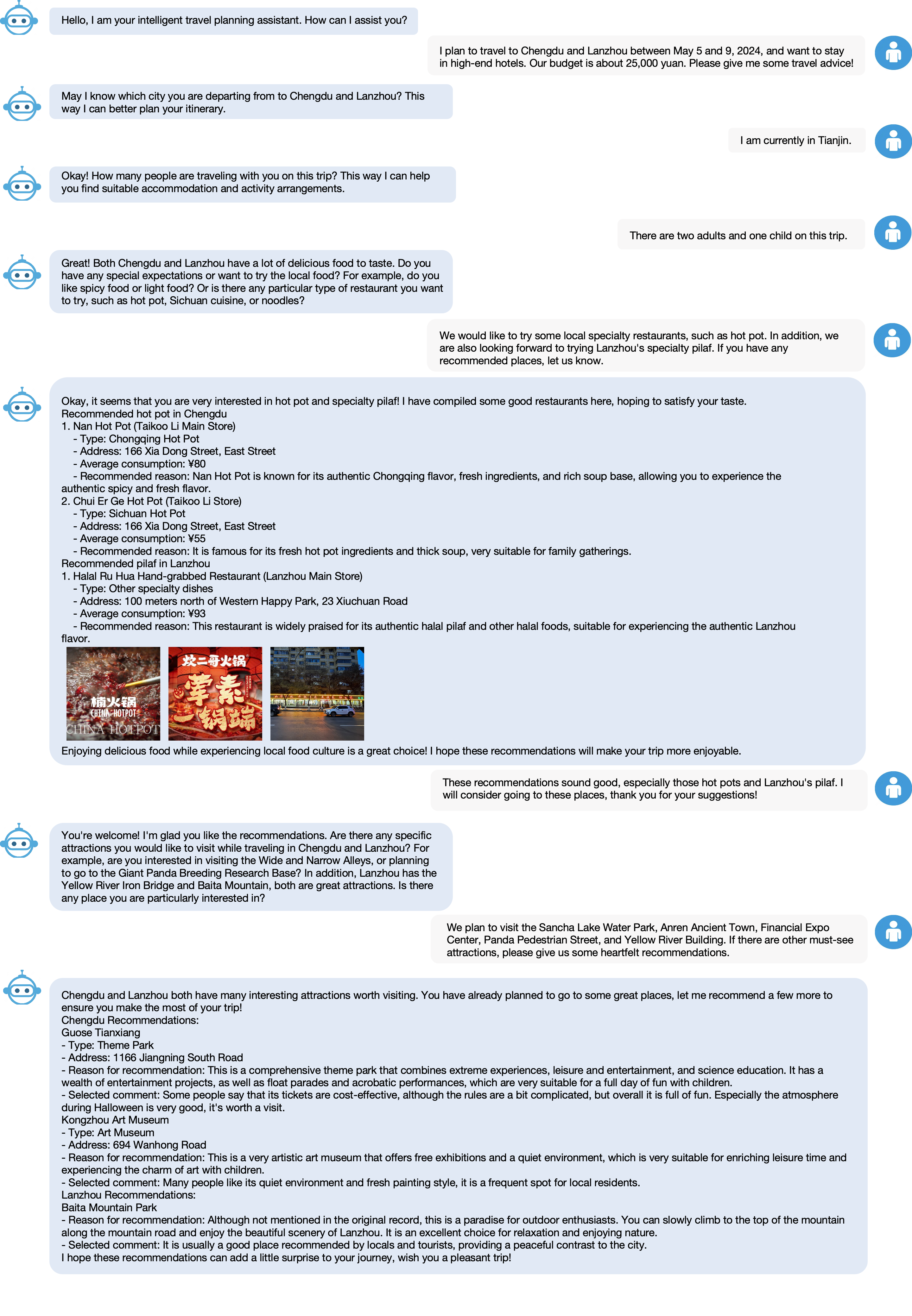}
\end{figure*}
\begin{figure*}[t]
\centering
\includegraphics[width=\textwidth]{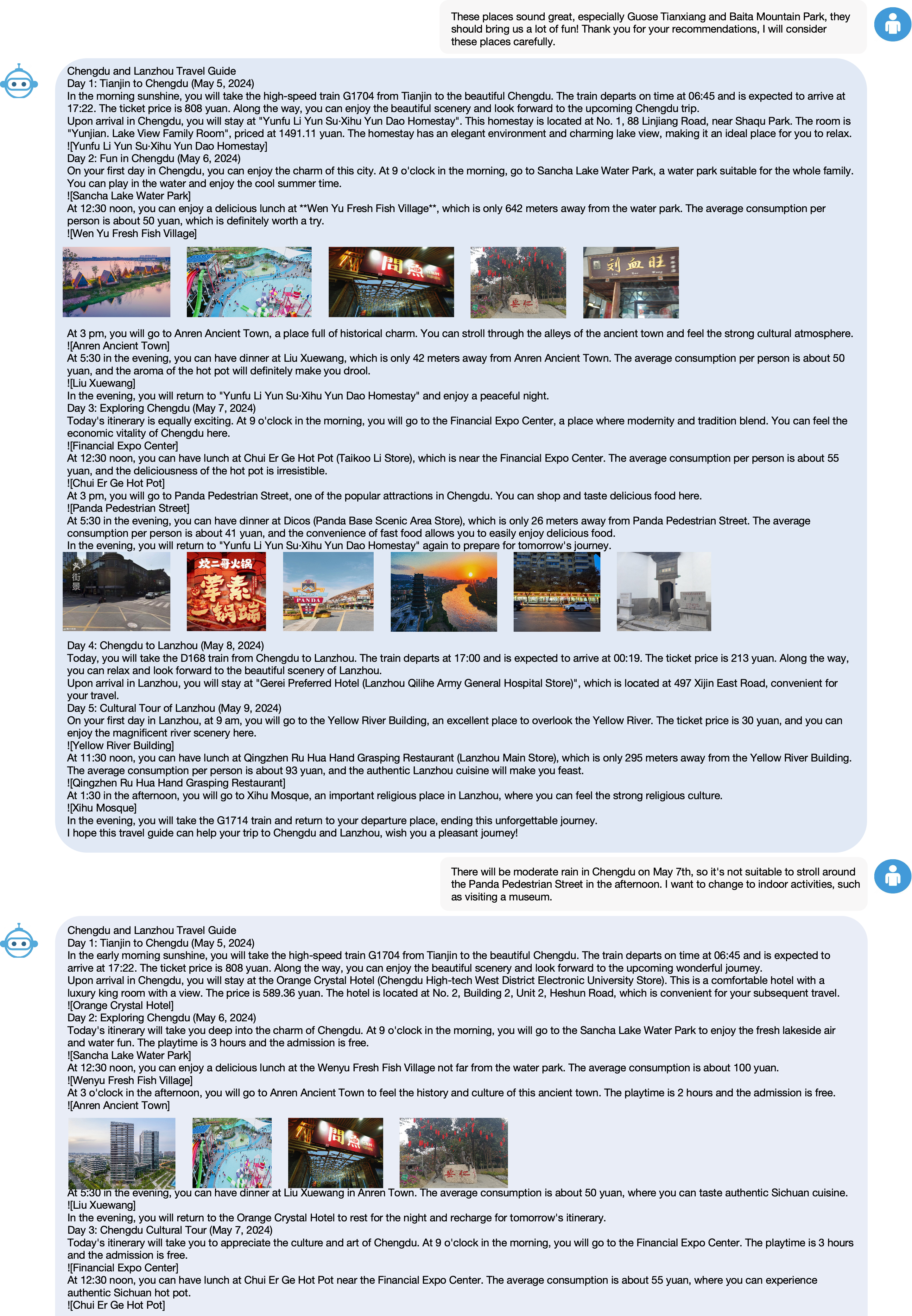}
\end{figure*}
\begin{figure*}[t]
\centering
\includegraphics[width=\textwidth]{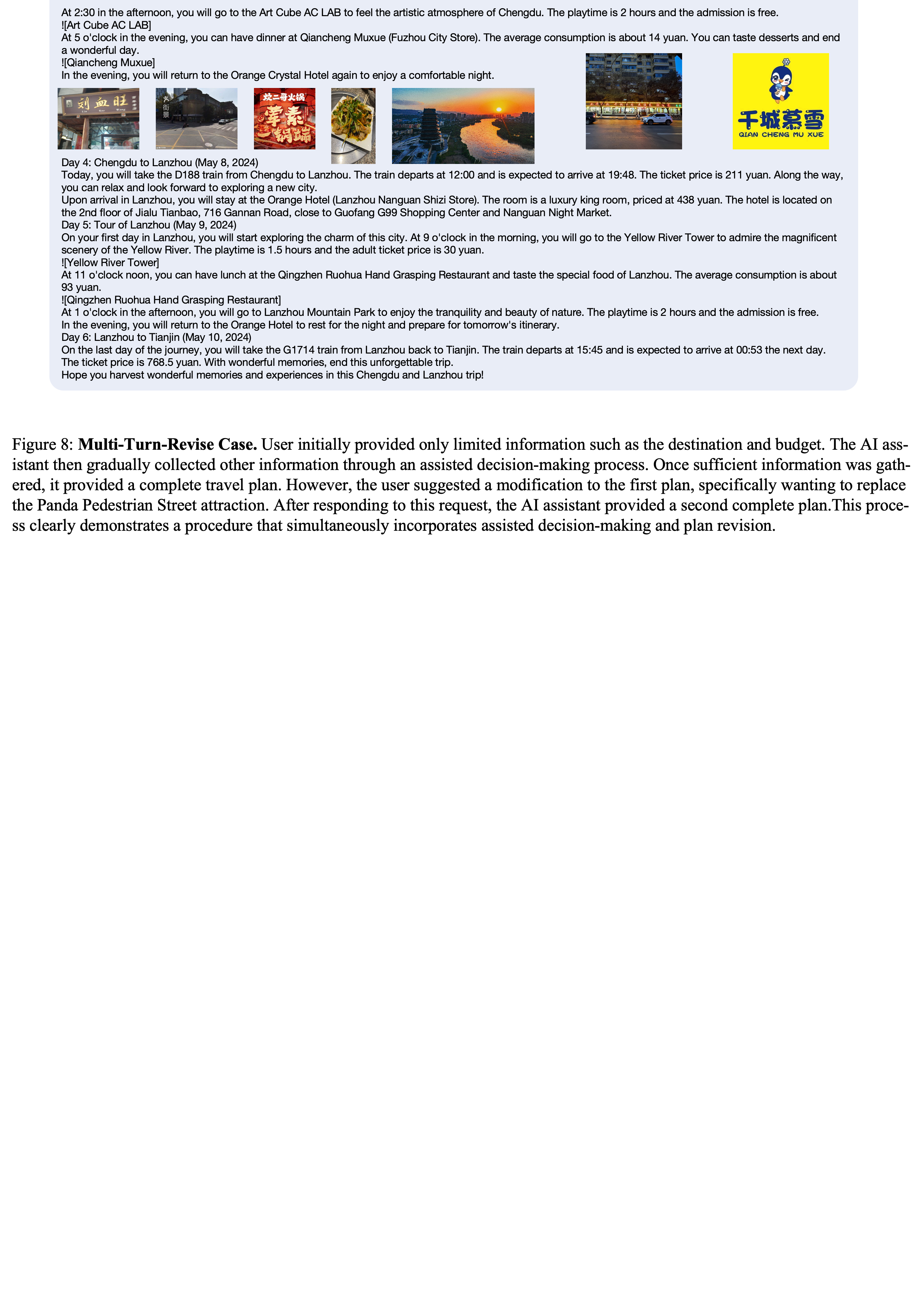}
\end{figure*}

\end{document}